\documentclass[a4paper,fleqn]{cas-sc}

\usepackage[authoryear,longnamesfirst]{natbib}

\def\tsc#1{\csdef{#1}{\textsc{\lowercase{#1}}\xspace}}
\tsc{WGM}
\tsc{QE}

\begin{document}

\shorttitle{Flowmind2Digital: The First Comprehensive Flowmind Recognition and Conversion Approach}

\title [mode = title]{Flowmind2Digital: The First Comprehensive Flowmind Recognition and Conversion Approach} 



%














\author[1]{Huanyu Liu\textsuperscript{1 }}
\author[1]{Jianfeng Cai\textsuperscript{1 }}
\author[1]{Tingjia Zhang\textsuperscript{1 }}

\author[2]{Hongsheng Li}
\author[2]{Siyuan Wang}
\author[2]{Guangming Zhu}
\author[3]{Syed Afaq Ali Shah}
\author[4]{Mohammed Bennamoun}
\author[2]{Liang Zhang\corref{cor1}}

\ead[url]{liangzhang@xidian.edu.cn}

\address[1]{School of Artificial Intelligence, Xidian University, China}

\address[2]{School of Computer Science and Technology, Xidian University, China}

\address[3]{School of Science and core member of Centre for AI and Machine Learning, ECU}
\address[4]{School of Physics, Maths and Computing, Computer Science and Software Engineering, UWA}

\cortext[cor1]{Corresponding author. \\ \textsuperscript{1} Equal contribution.}



\begin{abstract}
Flowcharts and mind maps, collectively referred to as flowmind, play an important role in our daily, and many companies have developed dedicated tools. Hand-drawn flowminds offer significant advantages for real-time and collaborative communication. However, there is an increasing demand to convert them into a digital format for further processing. Automated conversion methods are crucial in addressing the challenges associated with manual conversion, such as the cost of time and learning. Previous works have proposed diverse sketch recognition methods. However, these methods face significant limitations in practical situations. Firstly, most methods are designed for specific fields, making it challenging to extend their usage to other fields. Additionally, none of these methods address the critical step of digital conversion after recognition, which is essential to users. Moreover, existing datasets exhibit significant biases relative to actual data, hindering the methods' generalization ability.

~\

\noindent Our paper proposes the \emph{Flowmind2digital} method and \emph{hdFlowmind} dataset to address the aforementioned challenges. \emph{Flowmind2digital} is the first comprehensive recognition and conversion method for flowminds, utilizing a neural network architecture and keypoint detection technology to enhance overall recognition accuracy. Our \emph{hdFlowmind} dataset consists of 1,776 hand-drawn and manually annotated flowminds, covering 22 scenarios and surpassing existing datasets in size. Our experiments showcase the effectiveness of our method, with an accuracy rate of 87.3\% on the \emph{hdFlowmind} dataset, surpassing the previous state-of-the-art work by 11.9\%. Additionally, our dataset demonstrates effectiveness, with a 2.9\% increase in accuracy after pre-training and fine-tuning on Handwritten-diagram-dataset. We also highlight the importance of simple graphics for sketch recognition, which can improve accuracy by 9.3\%.
\end{abstract}


\maketitle

\section{\textbf{Introduction}}
\begin{figure}[h]
\centering
\includegraphics[width=3.5in]{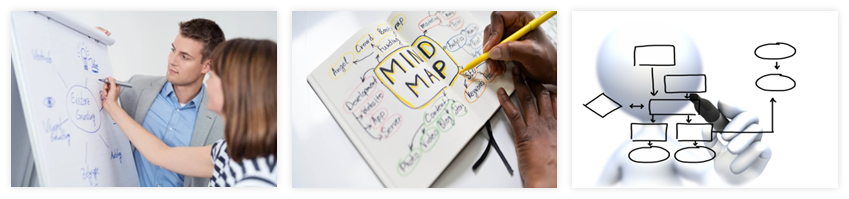}
\caption{Flowminds used in Various Scenarios}
\label{pho:img1}
\end{figure}

Sketching is a prevalent and innate communication skill in human society, dating back to ancient times and evident in the spontaneous drawings of infants. There are two practical forms of sketches: $(a)$ flowcharts, which visually represent the sequence of steps in a process and serve as an indispensable tool for documenting and revealing one's thought process, and $(b)$ mind maps, which are visual thinking tools that enable us to structure our ideas and create an intuitive framework around a central concept. Due to their usefulness, many companies have developed software for creating these sketches, collectively referred to as flowmind. However, flowminds are typically not created in any specific digital format but rather hand-drawn by users. These hand-drawn flowminds can be created during brainstorming and planning at a meeting, note-taking for academic research, or process planning as shown in Fig. \ref{pho:img1}. Hand-drawn flowminds have the advantage of real-time notation, as they only require a pen and any available background to immediately record structural information. However, interacting with software is not as direct, causing inconvenience in many scenarios.

However, there is a need to eventually convert these hand-drawn flowminds into digital format for the sake of clarity, better documentation, and record keeping. The process of converting hand-drawn flowminds into digital format involves a large amount of manual work on the software, including dragging each shape to the right position and typing textual labels for connectors (such as lines, arrows, and double arrows). This task requires considerable time and effort, especially for normal users who are not proficient in using such software.

To alleviate the burden of manual conversion, there is a need for an automated method that allows regular users to obtain digital flowminds directly on commonly used software from their hand-drawn input. This can be achieved in two steps: first, recognizing the elements in a hand-drawn flowmind; and second, converting the recognized elements into common formats using certain digital tools.

Existing methods have limitations and assumptions when it comes to recognizing elements in a hand-drawn flowmind, and they often produce output in a format that is not commonly used. For example, the UML model proposed by [\cite{gosala2021automatic}] only allows users to draw in predefined formats, which does not align with the initial sketch's intention. HDBPMN, proposed by [\cite{hdpmn}][\cite{schaeferSketch2BPMN2021}], offers a comprehensive approach that generates an XML file of a BPMN model from sketch input. However, this format is not commonly used in daily scenarios. Furthermore, no existing method has met the requirement for the final conversion to a commonly used format.

Existing datasets have limitations in terms of real-life scenarios, as they mainly consist of samples drawn on electronic devices. For instance, the Handwritten-diagram-dataset by [\cite{schaeferSketch2BPMN2021}] lacks diversity in terms of input sources. In contrast, real-life scenarios include photos of whiteboards or glass boards, which can present challenges such as overexposure or reflections, and are common sources of initial input. Therefore, there is a need for a comprehensive recognition and conversion method that can interface with commonly used software, such as Microsoft Power Point (PPT) and Visio, and a dataset that covers a larger scope of real-world scenarios.

Hence, our work makes two significant contributions. \textbf{\emph{Firstly}}, we present \emph{Flowmind2digital}, an approach that utilizes a neural network architecture based on the Arrow-RCNN model proposed by [\cite{arrowrcnn}] to automatically convert hand-drawn flowminds into editable PPT and Visio digital diagrams. We aim to improve the practicality of the existing approaches for normal users. \textbf{\emph{Secondly}}, we introduce \emph{hdFlowmind} dataset, which consists of 1,776 images and 27,804 annotations. This dataset covers a larger scope of scenarios and is considerably larger than previous datasets. Additionally, we include 485 samples of basic shapes in the dataset to demonstrate their importance in the experiments. We make this dataset publicly available for future research. Our experiments on this dataset indicate that Flowmind2digital outperforms the state-of-the-art models, which validates the effectiveness of hdFlowmind. Lastly, we also provide the first ultra-lightweight version of the Visio-Python kit that facilitates direct programming operations on Visio software.

Our work in this paper significantly extends the scope and quality of the Arrow-RCNN approach, which focuses on flowchart recognition and keypoint detection. Specifically, our contributions are as follows:

\begin{itemize}
\item We introduce an improvement based on the human keypoints detection that improves the accuracy of Arrow-RCNN.
\item We extend the model to cover the recognition of specific content of text box, which overcomes the limitation of Arrow-RCNN model that only recognizes the location of text.
\item We further achieve compatibility with the internal models of Detectron2\footnote{https://github.com/facebookresearch/detectron2.git}, providing a pre-trained model that demonstrates the usefulness of our \emph{hdFlowmind} pre-training approach for training in other tasks within the relevant sketch domain.
\end{itemize}

The rest of this paper is organized as follows: We provide an overview of existing methods and related work in the field of sketch recognition in Section 2, followed by several challenges that exist in the recognition scenario we focus on in Section 3. Section 4 provides a detailed description of our \emph{hdFlowmind} dataset. Then we introduce our \emph{Flowmind2digital} recognition model in Section 5 and evaluate our approach and dataset in Section 6. Finally, we discuss the implications and limitations of our approach in Section 7. The paper is concluded in Section 8.

\section{\textbf{Related Works}}
Drawing sketches is a useful method to convey information quickly and effectively. As a result, several sketch recognition approaches and datasets have been proposed to automate the conversion of sketches to digital formats. Microsoft's Visio and PowerPoint software have become widely used for creating digital diagrams. In this context, our work concentrates on flowmind recognition using keypoint detection and post-processing the results for use in PPT and Visio software. Therefore, we discuss the following related works in this paper: (1) existing flowmind recognition methods and datasets; (2) generic keypoint detection methods; (3) approach to interface with PPT and Visio software.

\subsection{\textbf{Flowmind Recognition}}
The process of flowmind recognition can be divided into three tasks including 1) shape recognition to locate and classify various basic shapes 2) connector recognition to identify the connector between each shape and the specific connection revealed using keypoints 3) text recognition to locate the text label and identify the specific content.

Conventionally, sketch recognition can be differentiated into online and offline methods which have diverse patterns. The input to online method is mainly a series of sequence strokes on geometry, stroke or gesture base. There are also datasets, including FC\_A proposed by Awal[\cite{awal2011first}] , FA and FC\_B proposed by Bresler et al.[\cite{bresler2016online}], DIDI [\cite{gervais2020didi}] proposed by Gervais et al. etc. [\cite{yu2003domain}][\cite{chen2008sumlow}]

In comparison to the online approaches, offline sketch recognition simply needs raw images instead of sequential strokes, making it more suitable for real world situations. However, this also brings greater complexity, given that less information is contained in the input. Some works [\cite{brieler2010model}][\cite{paulson2008paleosketch}][\cite{julca2020general}] tried to convert it to online problem by reconstructing the strokes of the hand-drawn flowcharts, but it is not applicable in the scenario we handle in this work which contains noises and raw images.

Therefore, the object-based method is also proposed, requiring corresponding datasets to train the models. Since the datasets of online and offline methods overlap, the online dataset can be converted into the offline datasets through certain transformation. Wu et al. [\cite{wu2015offline}] proposed to use FC\_A dataset to train offline recognition model [\cite{awal2011first}][\cite{costagliola2014local}]. Bresler extracted two datasets from the FC\_B dataset -- D\_a and D\_b [\cite{bresler2016online}][\cite{bresler2014recognition}][\cite{bresler2016recognizing}] for offline recognition.

At the same time, several datasets have been created to support sketch recognition for various scenes, including Bernhard's HDPBPMN dataset for BPMN graphs [\cite{schaeferSketch2BPMN2021}]. However, these datasets have limitations in terms of comprehensiveness, particularly in terms of exposure, drawing tools, and materials. These limitations will be discussed further in Section 3. Table \ref{tab:T1} provides an overview of the attributes of these datasets for different sketch scenes.

\begin{table*}[htbp]
\centering
\caption{Related Work and Datasets}
\scalebox{0.62}{
\begin{tabular}{cccccccccccccc}
\toprule
\multirow{2}[4]{*}{\textbf{Name}} & \multirow{2}[4]{*}{\textbf{Category}} & \multirow{2}[4]{*}{\textbf{Authors}} & \multicolumn{3}{c}{\textbf{Objects}} & \textbf{Connection} & \multicolumn{2}{c}{\textbf{Non-digital}} & \multicolumn{2}{c}{\textbf{Digital}} & \multirow{2}[4]{*}{\textbf{DB}} & \multirow{2}[4]{*}{\textbf{DPC}} & \multirow{2}[4]{*}{\textbf{DPT}} \\
\cmidrule{4-11} & & & Shapes & Connectors & Textboxes & Keypoints & Exposure & Blur & Pen-Pressure & DE & & & \\
\midrule
FC\_A[\cite{awal2011first}] & Flowcharts & {Awal et al.} & Yes & Yes & Yes & Yes & - & - & \textcolor[rgb]{ .5, .5, .5}{No} & \textcolor[rgb]{ .5, .5, .5}{No} & \textcolor[rgb]{ .5, .5, .5}{No} & \textcolor[rgb]{ .5, .5, .5}{No} &\textcolor[rgb]{ .5, .5, .5}{No} \\
\midrule
FA[\cite{bresler2014recognition}] & Finite automata & {Bresler,Pr\_u\_a} & Yes & Yes & Yes & Yes & - & - & \textcolor[rgb]{ .5, .5, .5}{No} & \textcolor[rgb]{ .5, .5, .5}{No} & \textcolor[rgb]{ .5, .5, .5}{No} & \textcolor[rgb]{ .5, .5, .5}{No} & \textcolor[rgb]{ .5, .5, .5}{No} \\
\midrule
FC\_B[\cite{bresler2016online}] & Flowcharts & {Bresler,Pr\_u\_a} & Yes & Yes & Yes & Yes & - & - & \textcolor[rgb]{ .5, .5, .5}{No} & \textcolor[rgb]{ .5, .5, .5}{No} & \textcolor[rgb]{ .5, .5, .5}{No} & \textcolor[rgb]{ .5, .5, .5}{No} & \textcolor[rgb]{ .5, .5, .5}{No} \\
\midrule
FC\_Bscan[\cite{bresler2016recognizing}] & Flowcharts & {Bresler,Pr\_u\_a} & Yes & Yes & Yes & Yes & - & - & \textcolor[rgb]{ .5, .5, .5}{No} & Yes & \textcolor[rgb]{ .5, .5, .5}{No} & \textcolor[rgb]{ .5, .5, .5}{No} & \textcolor[rgb]{ .5, .5, .5}{No} \\
\midrule
DIDItext[\cite{gervais2020didi}] & Flowcharts & {Philippe Gervais} & Yes & Yes & Yes & Yes & - & - & Yes & Yes & \textcolor[rgb]{ .5, .5, .5}{No} & Yes & Yes \\
\midrule
DIDIno\_text[\cite{gervais2020didi}] & Flowcharts & {Philippe Gervais} & Yes & Yes & \textcolor[rgb]{ .5, .5, .5}{No} & Yes & - & - & Yes & Yes & \textcolor[rgb]{ .5, .5, .5}{No} & Yes & Yes \\
\midrule
hdBPMN2021[\cite{schaeferSketch2BPMN2021}] & BPMN models & {Bernhard Schafer} & Yes & Yes & \textcolor[rgb]{ .5, .5, .5}{No} & Yes & Yes & Yes & \textcolor[rgb]{ .5, .5, .5}{No} & \textcolor[rgb]{ .5, .5, .5}{No} & Yes & Yes & Yes \\
\midrule
hdBPMN2022& BPMN models & {Bernhard Schafer} & Yes & Yes & Yes & Yes & Yes & Yes & \textcolor[rgb]{ .5, .5, .5}{No} & \textcolor[rgb]{ .5, .5, .5}{No} & Yes & Yes & Yes \\
\midrule
hdFlowmind & Flowcharts, Mind map & Ours & Yes & Yes & Yes & Yes & Yes & Yes & Yes & Yes & Yes & Yes & Yes \\
\midrule
\multicolumn{14}{c}{*DB = different backgrounds, DE = different equipment DPC = different pen colors, DPT = different pen thickness } \\
\bottomrule
\end{tabular}}%
\label{tab:T1}%
\end{table*}%

In terms of detection model, Julca-Aguilar and Hirata used the Faster-RCNN model [\cite{ren2015faster}] in target detection to identify flowchart elements [\cite{julca2018symbol}]. However, traditional object detection model can only tell where the arrows in a flowchart is, but not their correspondence (to where and what) . To address such limitation, Bernard et al. proposed Arrow-RCNN [\cite{arrowrcnn}], a model developed from Faster-RCNN, capable of detecting the head and tail of arrows in flowcharts, but lacked post-processing step to digitalize the result. Bernard et al. also proposed SketchToProcess [\cite{schafer2022sketch2process}], which realized the direct transformation of hand-drawn BPMN diagrams into corresponding standard BPMN xml files. However, the problem of connection orientation is still not solved. The scope of use is relatively narrow because of the limitations in recognizing the multiple connectors that commonly appear in mind mapping.

It is worth noting that most existing approaches do not provide a way to connect post-processing with previous software to generate editable graphics, which is a crucial requirement in practical applications. Most methods simply recognize and output raw data without further processing. The Sketch2Process model is an exception as it generates BPMN XML files, but even then, additional processing by users is still required. In practice, normal users require software-editable graphics rather than raw data that needs further steps. Therefore, the limitation of existing transformation models is evident from a practical perspective.

In summary, the analysis presented above highlights the active research field of sketch recognition for raw-sketch transformation, which has seen various approaches and datasets being introduced. However, there are still considerable gaps in the existing datasets in terms of their scope and quantity. Moreover, the current approaches do not consider the crucial step of connecting the recognized sketches to relevant software for practical use.

%

\subsection{\textbf{Keypoint Detection}}
Recognizing the connections in flowmind is a challenging task as it requires accurately identifying the trajectory of each connection. However, a more flexible approach is to simplify the target by recognizing a limited number of keypoints, which can improve the accuracy and flexibility of subsequent adjustments and generation in software.

Keypoint detection has numerous applications, one of which is human posture estimation (HPE) that includes the detection of facial, hand, and human keypoints. HPE systems such as those presented in [\cite{lin2017feature}][\cite{he2017mask}] typically follow a top-down approach for multi-keypoint detection. In the first stage, object detectors determine the location of each instance in an image. In the second stage, instances are clipped from the original image, adjusted to a specific resolution, and fed to an HPE network that outputs the keypoints' location. Currently, there are two main methods.

The first approach for keypoint detection involves using the keypoint location as the target for regression and directly outputting the location of the keypoint through a fully connected layer, without generating prediction results for each pixel. While this method has a lower model complexity and can achieve some end-to-end full differential training, it also has some obvious problems. These problems mainly include lower accuracy and generalization due to direct regression, no theoretical error lower bound, and a higher risk of overfitting because the task of keypoints is relatively flexible compared to HPE. Therefore, we do not use this method. The second approach is to estimate the heatmap based on keypoint location, which generates a heat value for each pixel position that indicates the probability of the corresponding pixel being a keypoint. Heatmap-based methods have been evaluated and have been shown to have higher accuracy in recent years, but there are still slight differences in the task of connector keypoints.

\begin{figure}[t]
\centering
\includegraphics[width=3.1in]{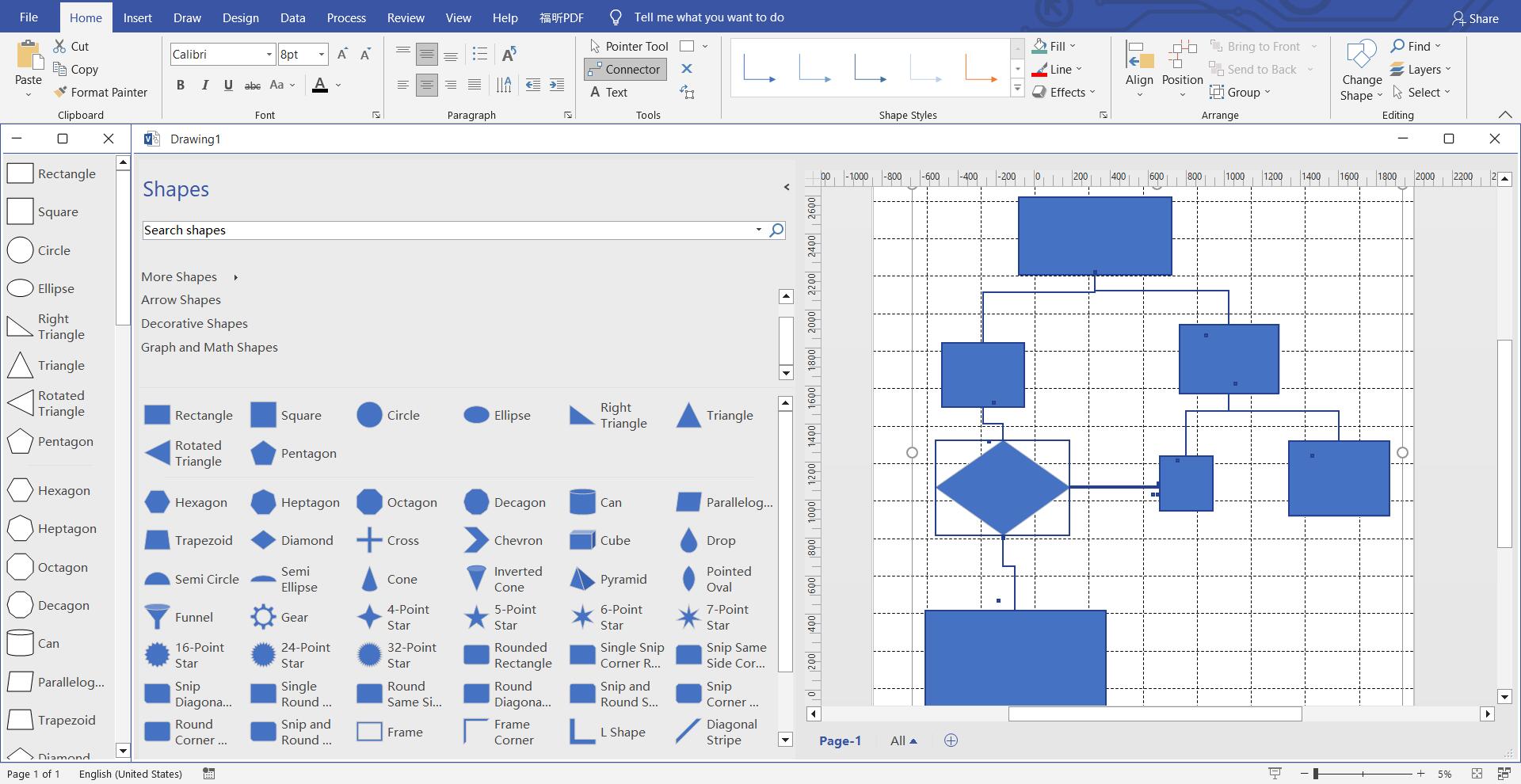}
\caption{Visio Interface}
\label{pho:img22}
\end{figure}

The traditional COCO dataset includes person instances that may be occluded, underexposed, or blurred, resulting in varying levels of visibility for the keypoints. The visibility is usually divided into three levels in general: "0" indicates that there is no keypoint. "1" indicates a invisible keypoint due to occlusion, and "2" indicates an existing visible keypoint. However, unlike keypoint detection for human posture estimation, where keypoints may have different levels of visibility due to occlusion, underexposure, or blur, keypoints of connectors typically have a clear and binary visibility state of either "0" or "2" since the head and tail of almost all arrows are clearly visible. Thus, the original keypoint detection method used for human posture estimation cannot be directly applied to connector keypoint detection. In this paper, we propose a modified model tailored to our specific needs, as discussed in Section 5.

\subsection{\textbf{Related Software}}
PowerPoint, a presentation developed by Microsoft Corporation, is widely used in flowchart creation and mind map editing. It has become a fundamental software tool in many industries. Visio is a graphics software tool in the Microsoft Office toolkit, which helps in understanding complex systems and processes and making better decisions. Visio diagrams can be saved in formats such as .svg and .dwg. As PowerPoint and Visio are widely used, our proposed solution aims to interface with these software tools. Fig. \ref{pho:img22} shows an example of the Visio interface.

Currently, the commonly used method for interacting with these software is to use the PPT and Visio API provided by Microsoft's official website\footnote{https://learn.microsoft.com/office/vba/api/overview/visio} \footnote{https://learn.microsoft.com/office/vba/api/overview/powerpoint}. This API offers detailed shape interfaces, internal functions, and proprietary properties, but it is primarily intended for developing their own extensions, making it difficult to interface with our model. Therefore, we have opted to use third-party extension packages to establish the interconnection.

The Python extension package "PPTX"\footnote{https://python-pptx.readthedocs.io} is used for interacting with PowerPoint software, and it adheres to the rules of PPT documents. The package defines pages as containers, shapes as objects, and connection lines as entities. Using this package, objects in PowerPoint can be manipulated through the Python language.

Regarding the Visio software, although there are third-party extension packages such as vsdx, their functions are too limited to meet basic needs. Therefore, we developed our API using the third-party extension package win32com provided by Python. This API mainly allows us to call the underlying components of Word, Excel, PPT, Visio, and other software.

\section{\textbf{Challenges}}
\begin{figure*}[t]
\centering
\includegraphics[width=6.5in]{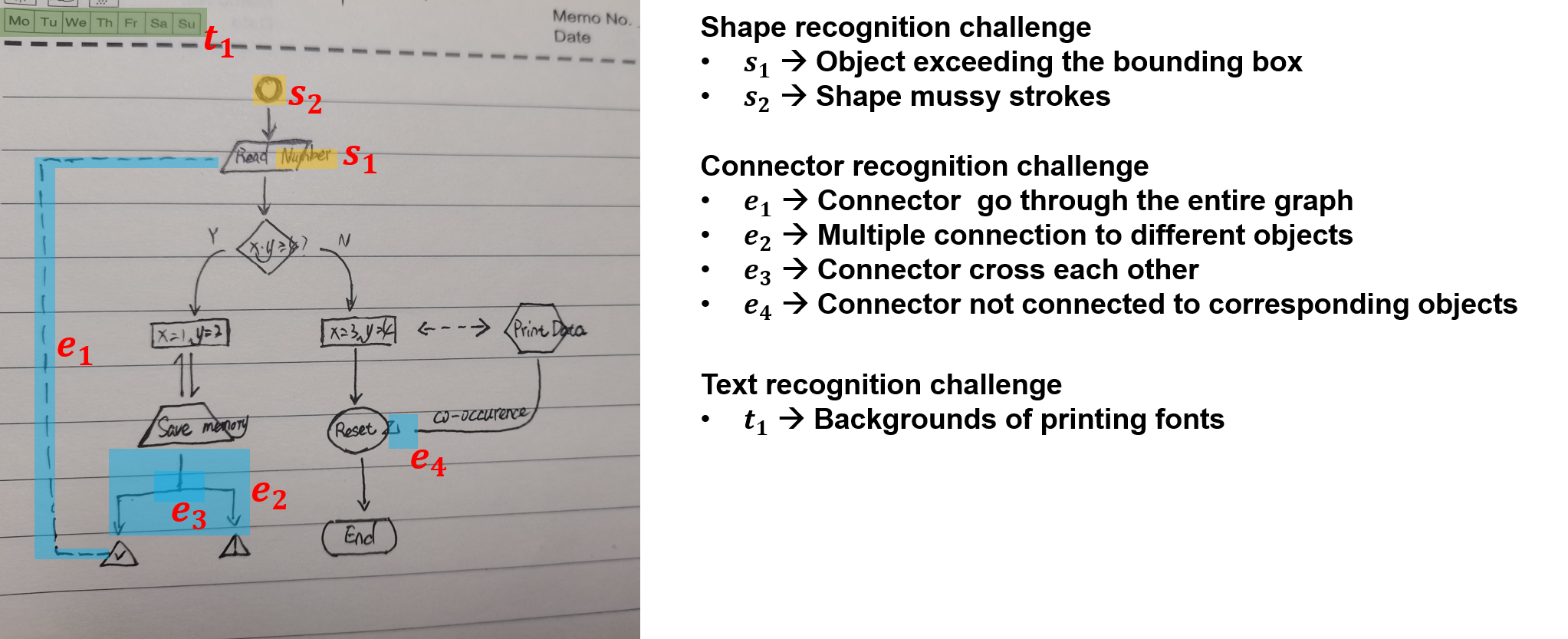}
\caption{Example of a flowmind with various highlighted recognition challenges}
\label{pho:img2}
\end{figure*}

This section discusses the difficulties in recognizing hand-drawn flowminds, which result from various real-world factors that are not adequately represented in existing datasets. We will focus on the challenges of shape recognition, connector recognition, text recognition, and the impact of the raw backgrounds. To illustrate these challenges, we will use the example drawing shown in Fig. \ref{pho:img2}, which is taken from our \emph{hdFlowmind} dataset.

\subsubsection*{\textbf{Shape Recognition Challenge}}
Recognizing all shapes and their positions in a flowmind is a major challenge in flowmind recognition, as it is the first step in the process. Shapes in a flowmind are referred to as nodes, and can include rectangles, diamonds, arrows, and text, each representing a different node. Shapes are defined by a bounding box, which contains information about the shape's location and type. However, recognizing shapes can be quite complex due to several challenges, including:
\begin{enumerate}
\item{One challenge in shape recognition is the ambiguity between similar shape types. For example, ellipses and long ellipses have a high degree of similarity, with the curvature of the edge being the only distinguishing factor (curvature is non-zero for ellipses, and zero for long ellipses). However, in practice, people tend to draw them similarly, which can lead to confusion in recognition.}
\item{The second challenge in shape recognition of hand-drawn flowminds is that the objects often exceed the bounding box of the shape, as shown in problem s1 in Fig. \ref{pho:img2}. This makes it difficult to distinguish between different shape types. Moreover, the boundary of the shape may be drawn multiple times, resulting in messy strokes, as depicted in problem s2 in Fig. \ref{pho:img2}.}
\end{enumerate}

\subsubsection*{\textbf{Connector Recognition Challenge}}
The connectors in flowcharts and mind maps play a crucial role in indicating the relationships between different nodes. Although many datasets classify connectors as a single type, they can actually be differentiated into three graphic styles, namely line, single arrow, and double arrow, which need to be recognized separately.

Recognizing connectors is more complex than recognizing shapes since connecting lines not only need to be identified in terms of their type and location, but also require the determination of the shapes they connect and the relationship between them. Some of the challenges associated with connector recognition are:

\begin{enumerate}
\item{The flexibility of the connection line path in hand-drawn flowminds can cause connectors to go through the entire graph, resulting in the bounding box including a large portion of other objects (e1 in Fig. \ref{pho:img2}).}
\item{Many-to-Many connections occur when a node extends multiple connection lines to different objects (e2), resulting in a high intersection over union (IOU) between the bounding boxes of these objects.}
\item{The crossing and intersection of connection lines, as shown in e3, can complicate the identification process since lines often cross over or intersect with other objects.}
\item{In rough sketches, bad connections are common where the drawn connectors are not always connected correctly to the corresponding object (e4). This issue makes it difficult to identify, especially when multiple possible objects are involved.}
\end{enumerate}

\subsubsection*{\textbf{Text Recognition Challenge}}
Recognition of text involves identifying both the textboxes and their respective positions, as well as the specific content within them. This task presents a few challenges such as:

\begin{enumerate}
\item{Handwriting recognition (HWR) is required to recognize the specific text content in the text box after locating it. However, in hand-drawn flowminds, the text often has more background interference compared to general HWR. For instance, people may draw their sketch on any background according to their convenience, and some backgrounds may even include noises such as printed fonts (t1).}
\item{Text that goes beyond the boundaries of a text box and intersects with other shapes or connectors is referred to as "Text out of text box". This can occur when the content of the text is too long and goes beyond the bounding box. Detecting this type of text presents similar challenges to those encountered when detecting shapes and connectors.}
\end{enumerate}
\begin{figure}[h]
\centering
\includegraphics[width=3in]{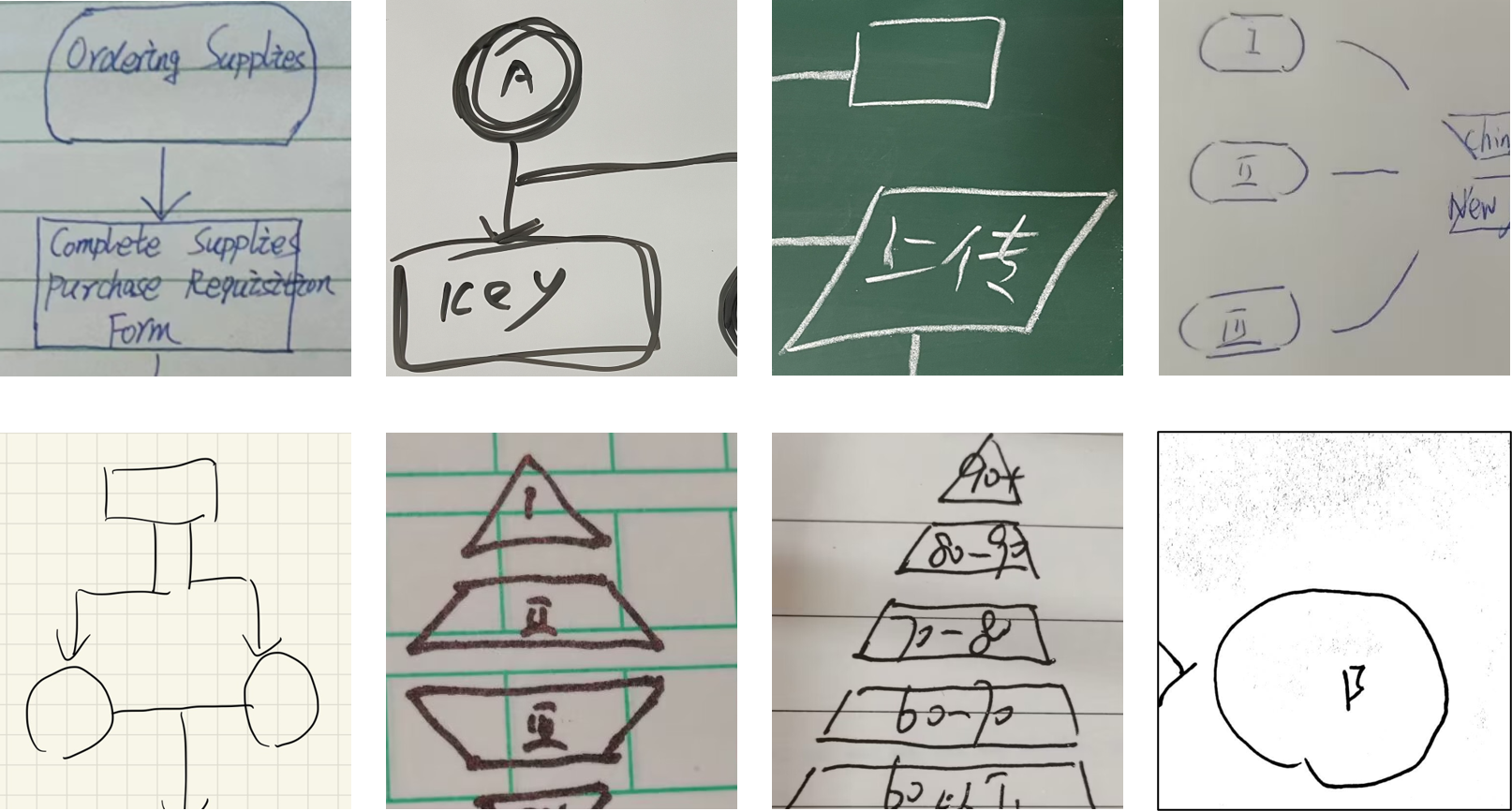}
\caption{Various Backgrounds in Flowminds}
\label{pho:img3}
\end{figure}
\subsubsection*{\textbf{Other Challenges}}
\begin{figure}[h]
\centering
\includegraphics[width=3in]{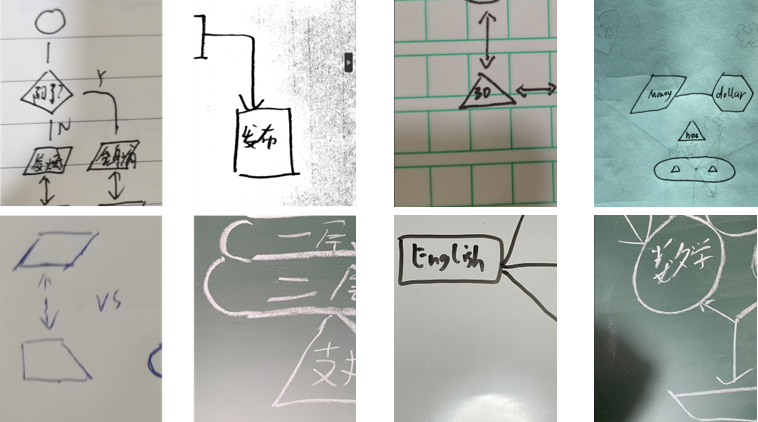}
\caption{Recognition Challenges Resulting from Different Methods of Digitizing Flowminds}
\label{pho:img4}
\end{figure}

Apart from the challenges mentioned earlier, there may be additional external factors that could impact the recognition process. The challenges in this case include:
\begin{enumerate}
\item
{The background of a hand-drawn flowmind can vary depending on the type of paper used, which may include horizontal lines, grids, squares, dashed lines, or blank spaces. The additional lines in the background may resemble the flowmind shapes or connectors drawn by the user, which can cause confusion during identification. Other scenarios may involve drawing on whiteboards, blackboards, or glass boards. Additionally, some users may use electronic devices for sketching, such as tablets or scanning software, which can result in variations in the background. Examples are shown in Fig. \ref{pho:img3}.
}
\item{The quality of the raw sketch can be influenced by the variability in drawing tools, such as the clarity, consistency, and thickness of the lines drawn.}
\item
{The equipment used to generate the input image can vary. Screenshots or scans taken directly on electronic devices typically have high clarity. However, taking a photo of a raw sketch on paper can result in rotated, blurry images or incomplete content. Examples illustrating this are shown in Fig. \ref{pho:img4}}.
\item{Another challenge lies in the post-processing stage after recognition. Existing methods do not consider the interface with software, leaving a gap in this aspect. We believe that we are the first to propose an interface with specific software for generating editable graphics}
\end{enumerate}

\section{\textbf{Dataset}}
This section discusses the Collection, Annotation, Characteristics and Splitting of the \emph{hdFlowmind} dataset, that is publicly available \footnote{https://huggingface.co/datasets/caijanfeng/hdflowmind}. Furthermore, emphasis is placed on the advantages of our dataset over relevant datasets in terms of quantity, quality, and diversity, showed in Table \ref{tab:T1}. This emphasis aims to comprehensively address and overcome the various challenges mentioned in the Section 3.
\subsection{\textbf{Dataset Collection}}
\begin{figure}[h]
\centering
\includegraphics[width=3in]{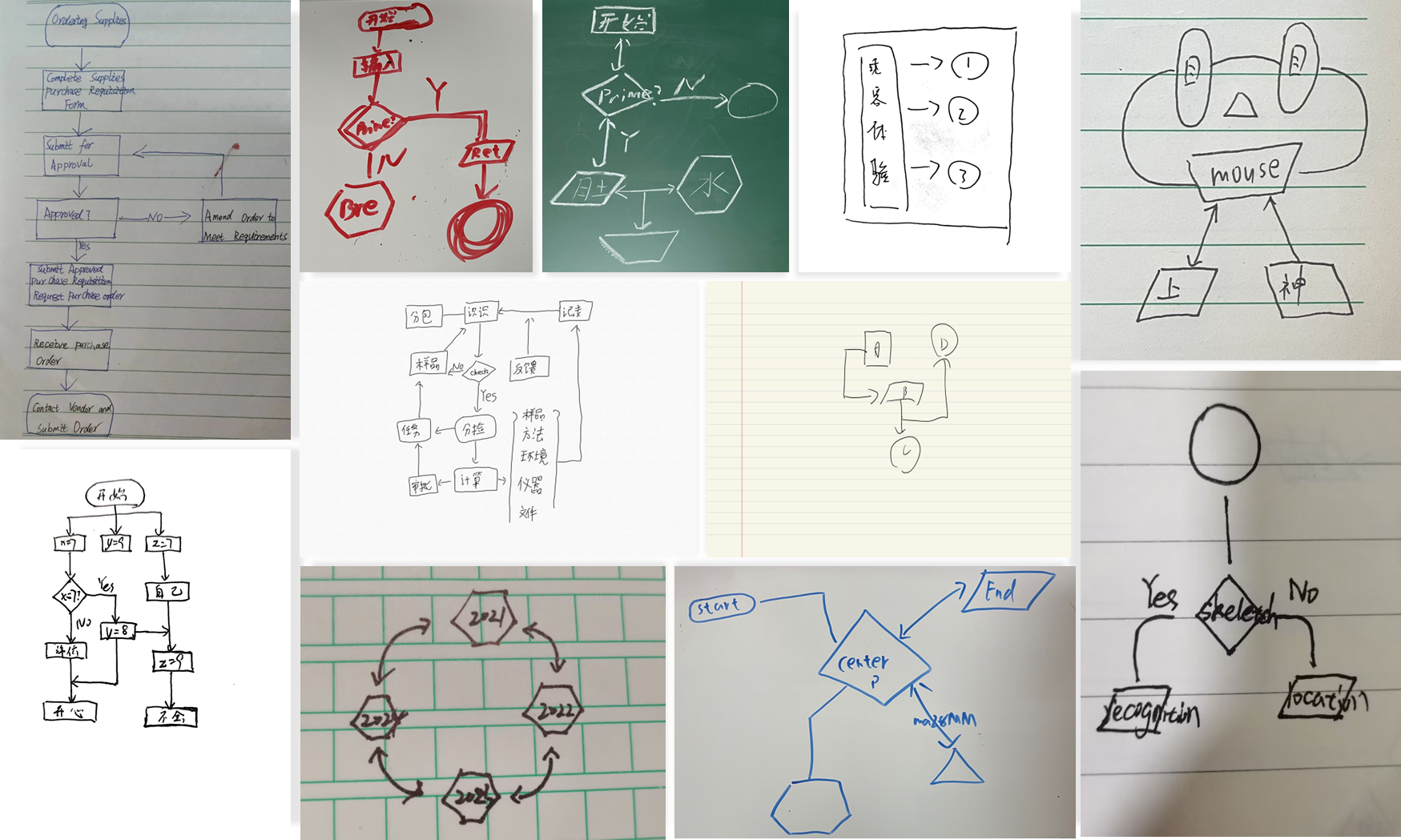}
\caption{Overview of the \emph{hdFlowmind} Dataset}
\label{pho:img5}
\end{figure}
Our dataset comprises 1,776 hand-drawn flowminds collected from XIDIAN University. These images include hand-drawn flowcharts found on the Internet, mind maps used in practical applications, as well as sketches made during meetings or brainstorming sessions. The participants were instructed to ensure that their designs and strokes reflect real-life scenarios.

To cover a larger scope of scenarios, we asked participants to draw with different drawing mediums, backgrounds, and photographic equipment. This intentional diversification in the dataset creation process ensures a more comprehensive representation of real-world conditions and user behaviors. Our dataset encompasses a rich variety of hand-drawn sketches captured under diverse conditions, including varying lighting, textures, and drawing styles. Unlike some existing datasets that may have limited diversity in terms of drawing styles or environmental conditions mentioned in Section 2, our dataset embraces a broader spectrum of user interactions and artistic expressions. This diversity in the data collection process enhances the generalization capability of our model, making it well-suited for real-world applications where visual inputs can vary significantly.

In the category of non-digital samples, We have a variety of non-digital samples in our dataset, which were drawn on different backgrounds such as whiteboards, glass boards, standard A4 paper, brown and yellow paper, thin paper (which is translucent and the content on the reverse side can be seen), grid paper, quadrille paper, lined paper (with line color in green or black), ruled paper, and chart paper (which has some patterns and text on it). The drawing tools used in these samples include markers, black pens, blue pens, whiteboard pens, chalks, and more. Additionally, the circumstantial conditions under which these samples were drawn include normal lighting, dusky lighting, overexposure, and shadow occlusion. We also have some samples that were scanned using software\footnote{https://www.camscanner.com/}.

For digital samples, participants used electronic devices such as Apple iPad and Samsung Pad, along with electronic pens, to create the sketches. They used two different software programs, Concept drawing board and Notability\footnote{https://concepts.app/} \footnote{https://notability.com/}. The backgrounds for these sketches included grids, lattices, horizontal lines, and scattered dots in various colors. The pen settings varied, including brush strokes with pressure sensing, fix jitter pen, and soft pencil tools provided in the software.

The overview of \emph{hdFlowmind} is provided in Fig \ref{pho:img5}.
\subsection{\textbf{Dataset Annotation}}
We utilized the PASCAL VOC format, which is a widely used format for Object Detection datasets in the field of Computer Vision, to annotate the shapes and connections in each sample for both training and evaluation purposes. To generate the annotation files, we utilized a data annotation tool provided by Huawei Cloud\footnote{https://cloud.huawei.com}. These annotation files consist of labels, four coordinates of bounding boxes, and additional coordinates of head and tail for connectors, which include arrow, double arrow, and line.

\begin{figure}[t]
\centering
\includegraphics[width=3in]{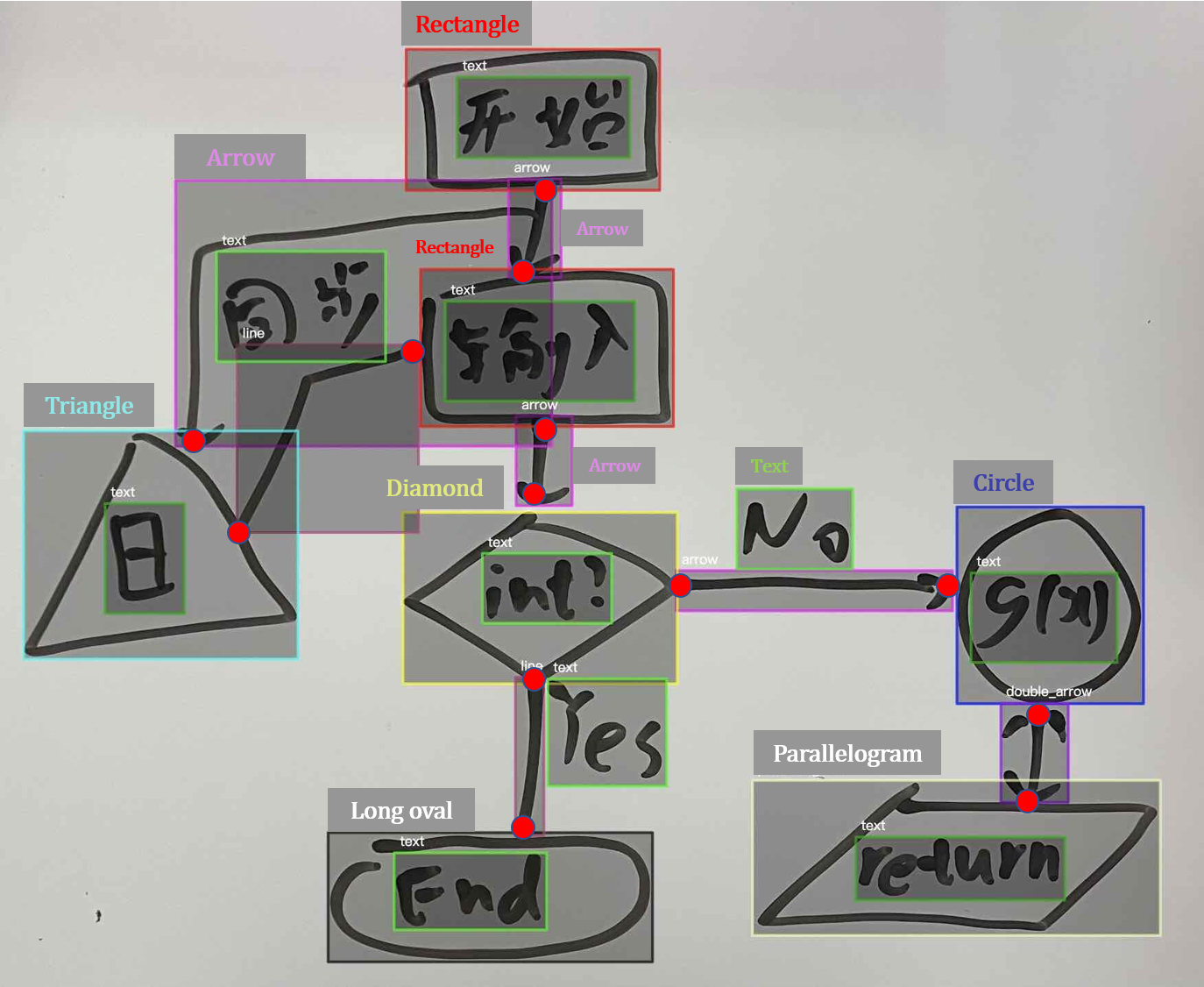}
\caption{An Example Image with Annotations}
\label{pho:img6}
\end{figure}

Please note that the annotation tool we used doesn't allow distinguishing whether keypoints are necessary for each instance. Hence, we used the coordinates of keypoints as attribute values of bounding boxes during annotation, and manually entered the coordinates afterward. After careful checking, the error rate was no more than 0.225\% (4/1776). Fig. \ref{pho:img6} shows a sample from our \emph{hdFlowmind} dataset.

\begin{table}[htbp]
\centering
\caption{\emph{hdFlowmind} elements in the 1,776 annotated images}
\setlength{\tabcolsep}{4.2mm}{
\begin{tabular}{ccc}
\toprule
\textbf{Element} & \multicolumn{1}{c}{\textbf{Name}} & \multicolumn{1}{c}{\textbf{Count}} \\
\midrule
\multirow{8}[16]{*}{Basic shapes} & \multicolumn{1}{c}{circle} & 1,039 \\
\cmidrule{2-3} & \multicolumn{1}{c}{diamonds} & 599 \\
\cmidrule{2-3} & \multicolumn{1}{c}{hexagon} & 562 \\
\cmidrule{2-3} & \multicolumn{1}{c}{long oval} & 485 \\
\cmidrule{2-3} & \multicolumn{1}{c}{parallelogram} & 700 \\
\cmidrule{2-3} & \multicolumn{1}{c}{rectangle} & 2,209 \\
\cmidrule{2-3} & \multicolumn{1}{c}{trapezoid} & 460 \\
\cmidrule{2-3} & \multicolumn{1}{c}{triangle} & 602 \\
\midrule
Text & \multicolumn{1}{c}{textblock} & 5,920 \\
\midrule
\multirow{3}[6]{*}{Connectors} & \multicolumn{1}{c}{arrow} & 3,219 \\
\cmidrule{2-3} & \multicolumn{1}{c}{double arrow} & 634 \\
\cmidrule{2-3} & \multicolumn{1}{c}{line} & 1,223 \\
\midrule
\multicolumn{3}{c}{Maximum of annotations per image: 93 } \\
\midrule
\multicolumn{3}{c}{Minimum of annotations per image: 3 } \\
\midrule
\multicolumn{3}{c}{Average of annotations per image: 22.56 } \\
\midrule
\multicolumn{3}{c}{Average of element per image: 14.23} \\
\bottomrule
\end{tabular}}%
\label{tab:T2}%
\end{table}%
\subsection{\textbf{Dataset Characteristics}}
Our \emph{hdFlowmind} dataset comprises 1,776 images, with 17,652 annotated bounding boxes and 10,152 keypoints. Table \ref{tab:T2} illustrates that the dataset covers 12 flowmind elements, including 7 basic shapes, 1 text box, and 3 types of connectors. The minimum number of annotations among hdFlowmind images is 3, while the maximum is 93, with an average annotation count of 22.56 per image. On average, each image contains 14.23 elements. Compared to similar datasets, hdFlowmind includes a larger variety of compositions and scenarios, with the most extensive quantity of images and annotations.

\begin{figure}[h]
\centering
\includegraphics[width=3.3in]{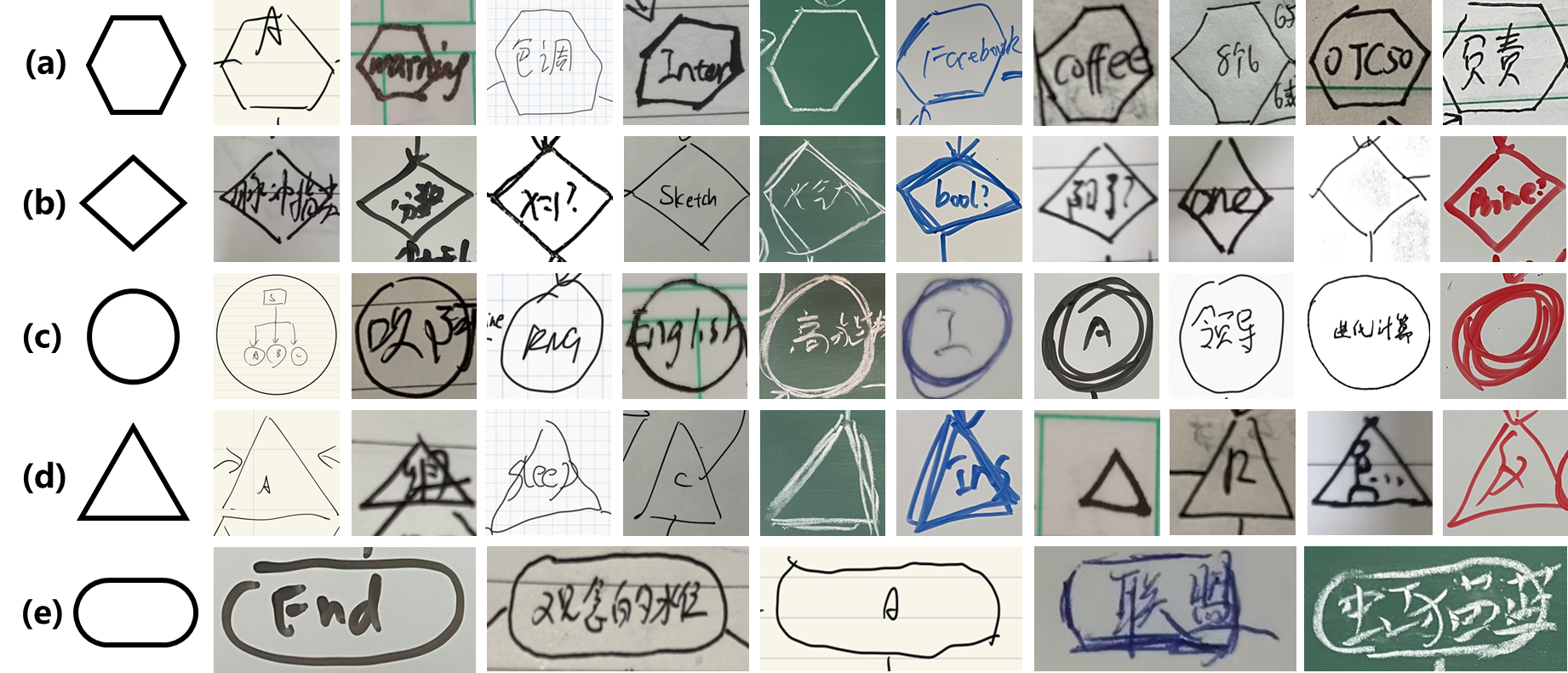}
\caption{Examples of different hand-drawn shapes}
\label{pho:img7}
\end{figure}

\begin{table}[htbp]
\centering
\caption{Statistics of challenges posed by different external features in our dataset}
\begin{tabular}{ccc}
\toprule
\textbf{Group} & \textbf{type} & \textbf{count} \\
\midrule
\multirow{8}[16]{*}{background} & whiteboard & 54 \\
\cmidrule{2-3} & blackboard & 48 \\
\cmidrule{2-3} & grid paper & 224 \\
\cmidrule{2-3} & ruled paper & 171 \\
\cmidrule{2-3} & brown paper & 200 \\
\cmidrule{2-3} & white paper & 155 \\
\cmidrule{2-3} & digital & 638 \\
\cmidrule{2-3} & single & 485 \\
\midrule
\multirow{6}[12]{*}{drawing tools} & marker & 54 \\
\cmidrule{2-3} & chalk & 48 \\
\cmidrule{2-3} & red pen & 4 \\
\cmidrule{2-3} & blue pen & 31 \\
\cmidrule{2-3} & black pen & 681 \\
\cmidrule{2-3} & electronic pen & 1,123 \\
\midrule
\multirow{2}[4]{*}{Other} & motion blur & 158 \\
\cmidrule{2-3} & exposure & 7 \\
\bottomrule
\end{tabular}%
\label{tab:T3}%
\end{table}%
The challenges in recognizing these images are mainly related to the issues discussed in Section 3, such as different backgrounds (paper, board, electronic devices, etc.) and drawing tools (markers, chalk, pens, etc.), as well as image-capturing problems (blurry images, exposure, etc.). Additionally, the recognition of shapes, connectors, and text also poses significant difficulties. These challenges are comprehensively presented in Table \ref{tab:T3}, while Fig. \ref{pho:img7} displays various types of shapes used in developing our dataset.

Our publicly available flowmind dataset has a wide range of elements and a high degree of composition diversity, which makes it valuable for research and development purposes. Additionally, our elements and compositions are practical and natural.

\subsection{\textbf{Dataset Splitting}}
\begin{figure}[h]
\centering
\includegraphics[width=3.3in]{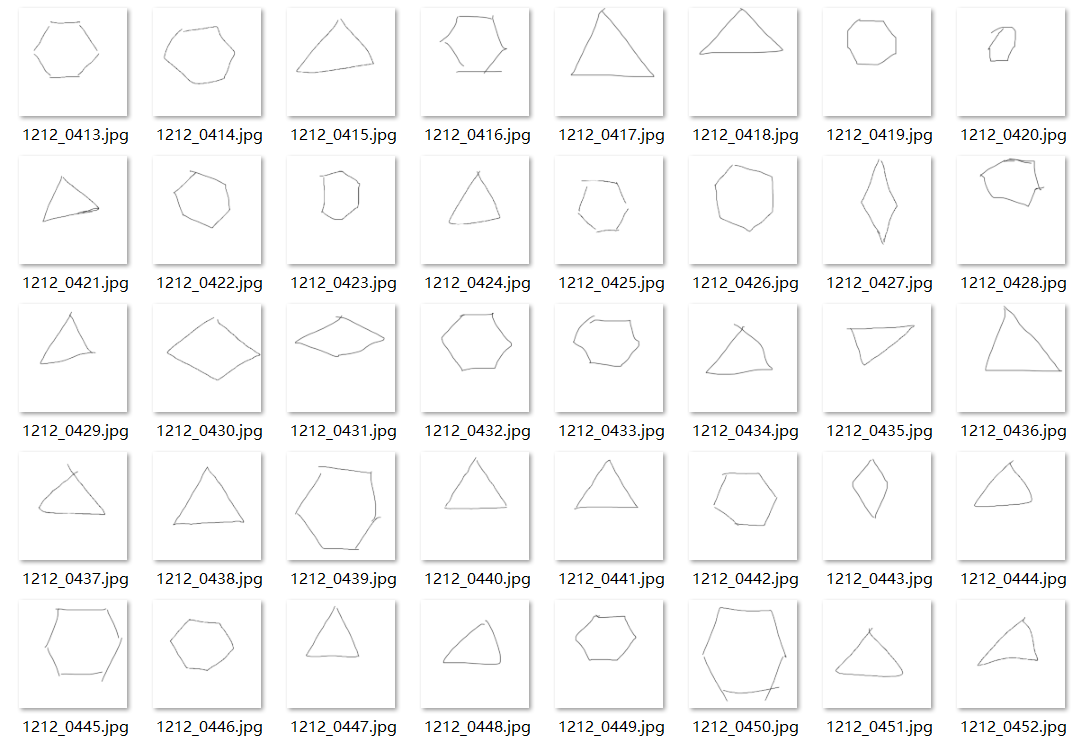}
\caption{Examples of Basic Geometric Shapes}
\label{pho:img8}
\end{figure}
To follow the protocol of related hand-drawn diagram datasets [\cite{bresler2016online}][\cite{bresler2016recognizing}][\cite{bresler2014recognition}], we split the \emph{hdFlowmind} dataset into three parts: training, validation, and testing. In comparison to professional field datasets, such as BPMN dataset [\cite{schafer2022sketch2process}], our flowmind dataset exhibits higher variability in terms of composition, backgrounds, and drawing medium. This implies that different flowminds from the same participant differ significantly. Thus, we randomly divided the dataset into three parts with a 6:2:2 ratio, i.e., the training set contains 775 samples, the validation set has 258 samples, and the test set has 258 samples.

f, to enhance the recognition accuracy of basic shapes with less frequency, we included an additional training image containing 107 parallelograms, 94 hexagons, 93 rhombuses, 102 trapezoids, 89 triangles, and a total of 485 basic images as depicted in Fig. \ref{pho:img8}. By adding this extra image, the ratio of train/validation/test sets became 1260:258:258. We assessed the impact of these auxiliary images on the training results in Section 6.

\begin{figure*}[h]
\centering
\includegraphics[width=6in]{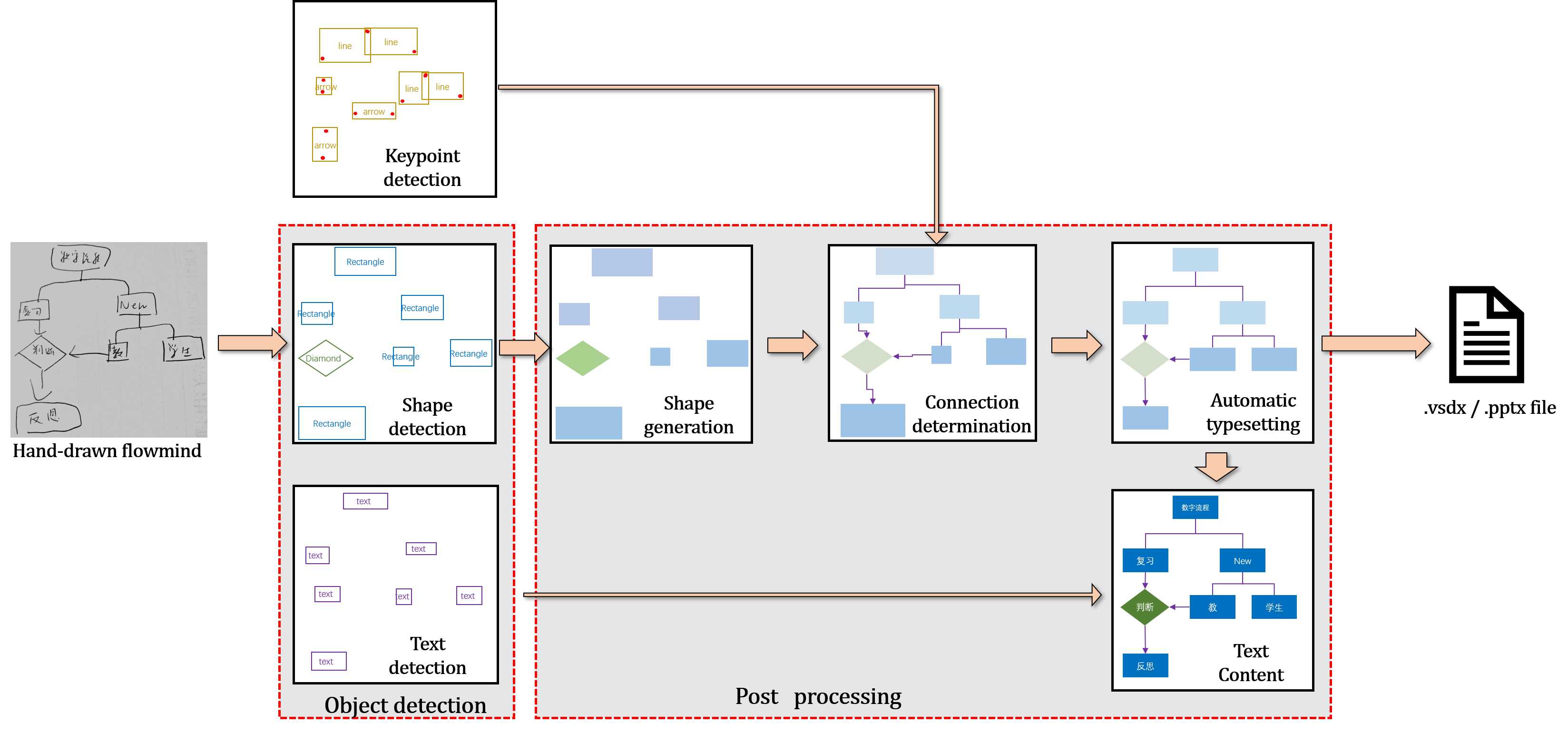}
\caption{Overview of our approach: Given a handwritten image, we first perform object detection for shapes, keypoints and text. Next, we generate the shape and connector based on coordinates and connection relationships in the relevant software. Then, we adopt a clustering-based automatic layout algorithm and fill in the corresponding text boxes detected by optical character recognition. Finally, we generate a visual file of the software.}
\label{pho:img9}
\end{figure*}
\section{\textbf{Methods}}
This section describes the \emph{Flowmind2digital} method for creating digital flowmind diagrams on PPT and Visio from hand-drawn inputs. As shown in Fig. \ref{pho:img9}, \emph{Flowmind2digital} consists of two main components: object \& keypoint detection, and post-processing.

\begin{figure}[t]
\centering
\includegraphics[width=3.5in]{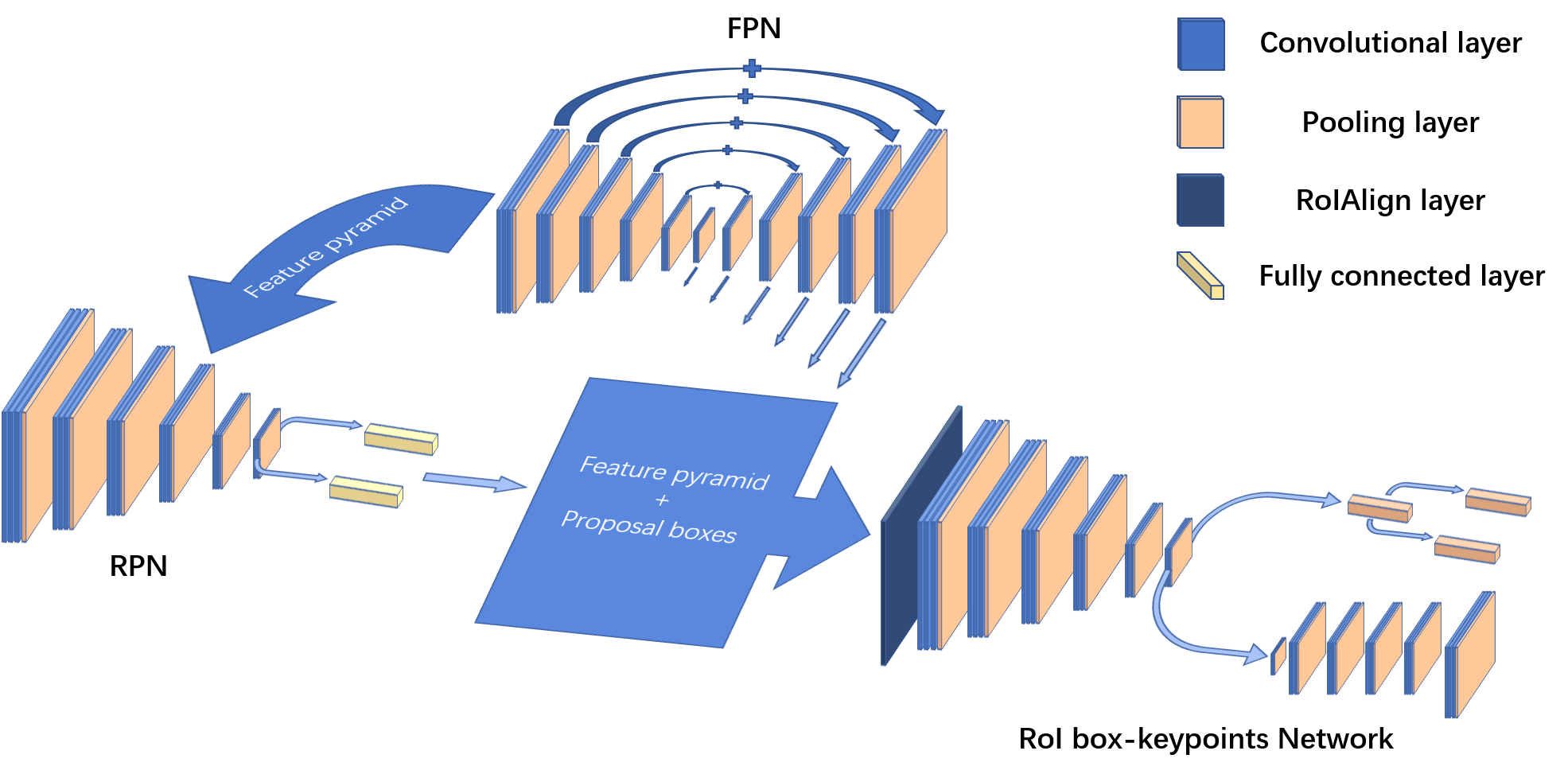}
\caption{Schematic of the Recognition Network in Our Model}
\label{pho:img10}
\end{figure}
\subsubsection*{\textbf{Keypoint detection}}
\subsection{\textbf{Object \& Keypoint Detection}}
\subsubsection*{\textbf{Mask-RCNN}}
The Mask-RCNN is a neural network approach based on the Faster-RCNN and is known for its high accuracy and extensibility, as described by [\cite{he2017mask}]. This approach can also be used for the keypoint detection, as demonstrated in the example of human posture recognition. The two-stage architecture, inherited from Faster-RCNN, provides a robust framework for effectively handling object detection tasks. This accuracy is crucial for tasks, where precise localization of keypoints is paramount. 

The initial input of Mask-RCNN is an RGB image represented as a three-dimensional array. The first two dimensions correspond to the size of the image, while the third dimension represents the three color channels: red, blue, and green. The detection process is based on the feature map of the image, which is learned by the backbone network, a FPN network in our application. Multi-scale feature is extracted and fused by up-sampling to combine the semantic and visual information.

The Mask-RCNN follows a two-stage process, similar to Faster-RCNN, for object detection. In the first stage, the Region Proposal Network (RPN) generates region proposals. Each proposal represents a bounding box around a region of interest along with a confidence score that indicates whether it belongs to the foreground or background. The second stage of Mask-RCNN involves a ROI (Region of Interest) network that classifies each proposal and refines the bounding box location. The ROI Align mechanism is used to extract a fixed regional feature map that corresponds precisely to the proposal region. This smaller feature map is used to predict refined bounding boxes and a score distribution over defined classes for each foreground proposal. Finally the Mask-RCNN gives an output set of object $(x,c,s)$, where $x$ represents coordinates, $c$ represents the most likely class and $s$ represents the confidence score.

Moreover, the ability of Mask-RCNN to perform keypoint detection aligns well with the specific requirements of the application. The mechanism for region proposal generation in the first stage, coupled with the ROI network in the second stage, enables not only accurate bounding box localization but also detailed keypoint predictions.

According to[\cite{amjoud2023object}][\cite{zou2023object}], we will conduct a comprehensive analysis of trade-offs in the entire system pipeline, considering aspects such as program runtime, memory usage, accuracy, speed, and correctness. Detailed experiments are presented in Section 6 to provide a thorough examination of the rationality behind the utilization of Mask RCNN.

\textbf{Trade-off in Speed and Time Complexity}

While Mask-RCNN may not be the fastest algorithm, the trade-off in speed is justified by the enhanced accuracy achieved through the two-stage process. The use of the Region of Interest (ROI) Align mechanism contributes to a time complexity suitable for our application. Comparatively, YOLO[\cite{redmon2016you}] may offer faster inference times but at the expense of potential sacrifices in accuracy.

\textbf{Memory Efficiency and Multi-scale Feature Fusion}

The FPN backbone utilized in Mask-RCNN contributes to effective memory utilization, accommodating the processing of high-resolution images. The multi-scale feature extraction and fusion through up-sampling enhance the model's ability to capture both semantic and visual information, a feature not explicitly highlighted in some other architectures. DETR[\cite{carion2020end}], while novel in its transformer-based architecture, may have different time complexity considerations as it directly predicts object bounding boxes and classes in a single pass. Due to the parallelization mechanism of multi-heads in the DETR transformer architecture being optimized for GPUs, it cannot run efficiently on lightweight CPUs.

Over all, based on previous research work and analysis above, we utilize the Mask-RCNN for our object detection network.

Keypoint detection is a popular extension of the Mask-RCNN model. During the second stage, the ROI network is altered to extract keypoints and refine bounding boxes at the same time. The ROI Box-Keypoints network involves two parallel fully connected layers, which are applied after the feature map array is transformed. One of the layers is responsible for refining the bounding box by regression, while the other layer treats keypoint detection as a pixel-level classification problem. Each arrow instance is assigned a one-hot mask, with the keypoint pixel defined as foreground and the rest as background.

We utilized a pre-trained FPN-resnet50 model, which was originally developed for human posture recognition. The selection of FPN-resnet50 as the backbone is motivated by its well-established balance between high accuracy and robustness. Additionally, it facilitates seamless experimental comparisons with other models of a similar class. But we customized it for our purpose. \textbf{Specifically, we set the number of keypoints to be detected as two.} This was the basic number needed to ensure that the detector can identify the connection between two shapes. The path of the connector can be easily adjusted through automatic typesetting or manual post-processing, that we need no additional keypoints to indicate the specific path of the connector. Therefore, two keypoints are sufficient to capture the composition of the flowmind diagram.

\subsection{\textbf{Post processing}}
After obtaining the results of identifying key points for objects and connectors as described earlier, our attention turns to the human-computer interaction process. We aim to create a technique that enables the conversion of these results into a digital format. This involves generating files such as .pptx or .vsdx from the output. 

Our approach to accomplishing this task involves the creation of a post-processing procedure, as described in Section 2. This procedure can be broken down into four distinct parts.

Firstly, the shape generation component uses Visio and PPT to create the corresponding shapes at the coordinates detected during keypoint recognition.

Secondly, the connection determination component identifies the exact point on the shape where the detected connector should connect, based on the keypoint information.

Thirdly, the text content component assigns labels to the shapes and connectors, generates text boxes, and extracts the relevant content using OCR software.

Lastly, the automatic typesetting component adjusts the sizes and positions of the shapes based on intelligent clustering, and generates the final output.

\subsubsection*{\textbf{Shape generation}}
As discussed in Section 2, we utilize the python-pptx library to establish communication with Microsoft Powerpoint. This library provides Python classes to interact with the elements present in a PPT document such as slides, shapes, and connectors. In the case of Visio, since there is no suitable toolkit available, we use the win32com\footnote{https://pypi.org/project/pywin32/} library to interact with the program. The Visio template is read and placed onto the created page object.

As mentioned in Section 2, we use the python-pptx library to interface with Microsoft Powerpoint. The elements of slides, shapes or connectors in ppt document is operated as python classes. Since there's no proper toolkit to interface with Visio, we use the win32com to operate it with this program. The template in Visio is read and placed onto the created page object.
To ensure proper formatting, we need to convert the coordinates of the bounding box $ (x_{0}, y_{0}, x_{1}, y_{1}) $ to the format of $ (x_{c}, y_{c}, H, W) $ using the following formula:

\begin{align}
\left(x_{0}, y_{0}, x_{1}, y_{1}\right) \rightarrow\left(x_{c}, y_{c}, H, W\right) \quad \quad \quad \qquad
\text { s.t. }\left\{\begin{array}{l}
x_{c}=\frac{\left|x_{0}+x_{1}\right|}{2} \\
\\
y_{c}=\frac{\left|y_{0}+y_{1}\right|}{2} \\
\\
W=\left|x_{0}-x_{1}\right| \\
\\
H=\left|y_{0}-y_{1}\right|
\\
\end{array}\right.
\end{align}

\subsubsection*{\textbf{Connection determination}}
\begin{figure}[h]
\centering
\includegraphics[width=3.3in]{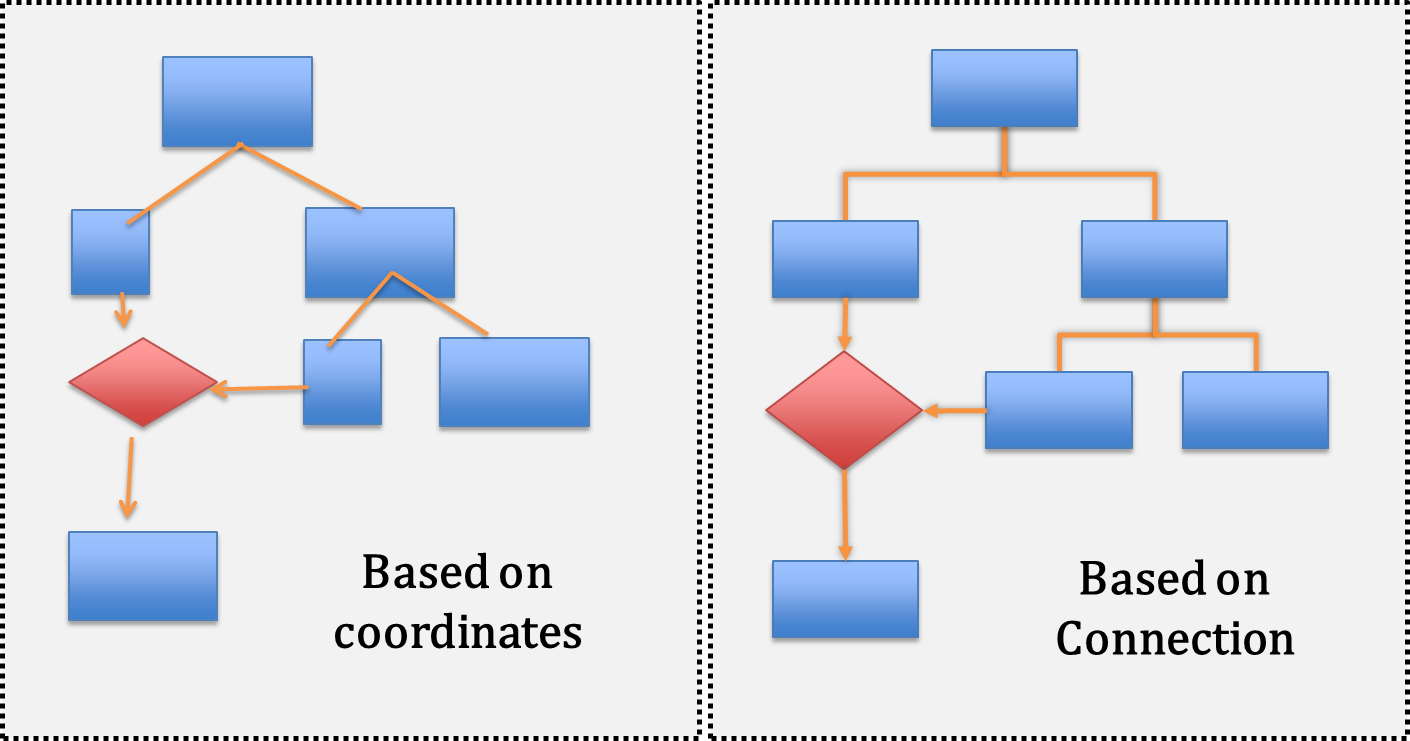}
\caption{The difference of two generation methods based
on coordinates and connection}
\label{pho:img11}
\end{figure}

After generating the Flowmind components, the next task is to link the connectors to the keypoints on the shape. In both Visio and PPT, if a connector is generated solely based on coordinates, it will remain fixed in its original position even if the shape is moved. To establish a true relationship between the shape and the connector, it is essential to connect it with the pre-defined keypoints on the shape. Figure \ref{pho:img11} illustrates the difference between the two methods of connector generation based on coordinates and connection.

\begin{figure}[h]
\centering
\includegraphics[width=3.3in]{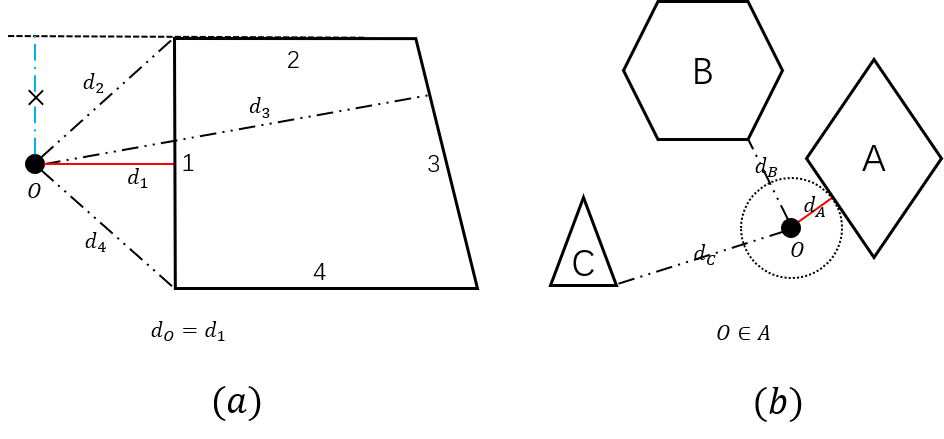}
\caption{Calculation of distance between connectors and different types of geometry shapes}
\label{pho:img12}
\end{figure}

Our algorithm aims to establish a connection between connectors and shapes by calculating the Euclidean distance between their keypoints. Euclidean distance serves as an intuitive and effective metric, accurately expressing the distance between keypoints of connectors and various geometric shapes. Its application facilitates the calculation of distances between connectors and candidate points on shapes, derived from the geometrical relationship between bounding boxes and standardized shapes. This approach, which includes calculating distances for polygons, identifying candidate points for non-polygons, and selecting the nearest shape for each keypoint, demonstrates robustness and adaptability in handling diverse geometric scenarios. The flexibility of Euclidean distance enhances the algorithm's ability to accurately model and understand geometric relationships, ensuring a robust and reliable approach for connector and shape connection in the recognition of complex hand-drawn sketches. The distance calculation between connectors and various types of geometric shapes can be derived using the geometrical relationship between the bounding box and the standardized shape, as shown in Fig. \ref{pho:img12}.

Assuming that the detected shapes in bounding box is set in an standardized orientation (with a horizontal base), the process first calculate the candidate points on each shape, referring to the connectable anchors on PPT and Visio shapes. Secondly for each connector keypoint, it identifies the nearest candidate point on all shapes. As for polygons, it computes the vertical distance from the keypoint to each edge (as depicted in Fig. \ref{pho:img12}a $d_1, d_3$). Note that, if the foot point of that vertical line lies on the extension of the edge, it chooses the shortest distance from the keypoint to terminal point of the edge as the shortest distance instead (as depicted in Fig. \ref{pho:img12}a $d_2, d_4$). For non-polygons, it identifies $n$ candidate points on the shape (according to the connection rules of PPT and Visio) and specifies that the keypoint can only be connected to them. For instance, the candidate points of a circle are Up, Down, Left, Right, Top left, Bottom left, Top right, and Bottom right.
Lastly, for each keypoint, the shape with the nearest candidate point is selected as the connected object (as depicted in Fig. \ref{pho:img12}b).


\subsubsection*{\textbf{Text content}}
In this step, our approach aims to recognize specific content within the text boxes. The input consists of a set of coordinates for each textbox, T, and the image feature map, F. The output includes the recognized content set, C, and a confidence score, S. Instead of training an OCR model from scratch, we use an existing OCR model, which offers superior accuracy and speed. The OCR model is also a two-stage method that generates bounding boxes in regions of interest and then recognizes the specific content of each bounding box. However, due to the challenges mentioned in Section 3, its performance on hand-drawn flowminds is considerably limited. Therefore, we apply OCR in each text box identified previously, which greatly improves the accuracy of text recognition, as demonstrated in Section 6.

To address merging or splitting problems in text box recognition, we utilized the method proposed in Arrow-RCNN [\cite{arrowrcnn}] to create a unified text box with a union bounding box that covers the corresponding text boxes. To determine which shape or connector a textbox corresponds to, we calculated the intersection over Union (IoU) between each text box and all bounding boxes of shapes. As mentioned in the shape recognition challenge, an IoU threshold of 80\% was set. If there exists a shape that has the highest IoU rate over the threshold for a detected text box, its content is filled into that shape. Otherwise, it is considered as an independent text element. As connectors have high flexibility, we created the text box through the corresponding bounding box to fill in its content.

\subsubsection*{\textbf{Automatic typesetting}}
\begin{figure}[h]
\centering
\includegraphics[width=3.3in]{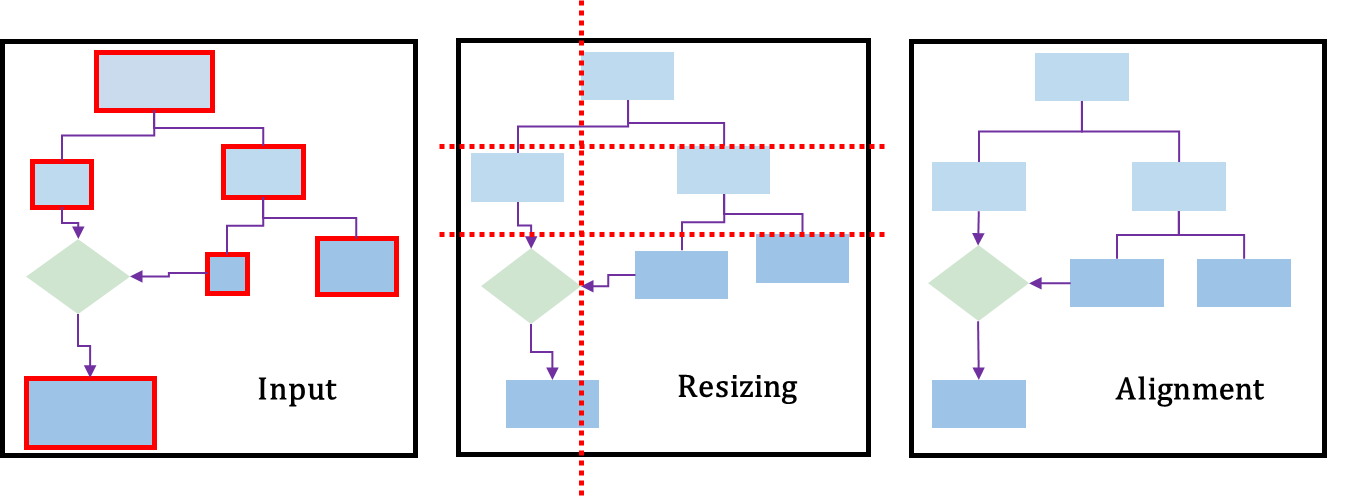}
\caption{Automatic Typesetting: Intelligent Scaling of Shape Sizes, Followed by Automatic Horizontal and Vertical Alignment}
\label{pho:img13}
\end{figure}

The steps described earlier have established the inclusion and graph relations between shapes, connectors, and text boxes, primarily realizing the visualization. However, a precise copy of the rough sketch may not always reflect the user's intention. For instance, the rectangles in Fig. \ref{pho:img13} are meant to be of the same size, but due to the visualization being based on the shape of the bounding box, there may be slight variations in the actual digitization. Additionally, the rectangles should be vertically aligned, but the digitized coordinates may not reflect this due to differences in the actual sketch.

The aforementioned deviation can pose problems, especially when creating a digital flowchart automatically. In a manually created flowchart, the consistency of a set of shapes can be ensured by copying and pasting. However, if the flowchart is already generated on the software, adjustments to individual shapes have to be made separately as deleting or replacing any shape can affect the established relations. Hence, an automatic typesetting algorithm that can assist in intelligent typesetting becomes particularly important.

To achieve this, we have implemented a two-stage clustering model that employs the Canopy and K-means algorithms. The number of clusters is determined through Canopy clustering, which is then followed by K-means to produce the final result. Moreover, we have utilized these clustering algorithms to adjust the size of shapes to account for variations in the input flowminds. Generally, the clustering algorithm is applied twice for automatic resizing and alignment. A summary of the clustering algorithm is depicted in Fig. \ref{pho:img13}.

To resize the editable graphics, we utilize a two-stage clustering model based on Canopy [cite] and K-means algorithm. Firstly, we use the length and width of the bounding box as clustering features. We set the thresholds of the Canopy algorithm and consider shapes with similar length and width as a cluster in coarse clustering. This provides us with a clustering reference value K. In the next stage, we perform fine-grained clustering using K-means algorithm, and calculate the average size of the bounding box for each cluster. This average size is used as the new size for the shape cluster. We perform this resizing process twice, to ensure automatic resizing and alignment of the shapes.

Moreover, Canopy clustering stands out by eliminating the need for a pre-specified k value, making it exceptionally practical. Despite potentially lower accuracy compared to other clustering methods, Canopy excels in speed, making it a valuable choice. Hence, Canopy clustering is strategically applied for preliminary coarse clustering, allowing the machine to autonomously determine the $K$ value and approximate $K$ initial centroids. Subsequently, this is followed by a more detailed fine clustering using K-means. The Canopy+K-means clustering strategy not only balances speed and accuracy but also proves effective in the context of automatic typesetting, emphasizing the machine's ability to specify the number of clusters in subsequent work.


To achieve alignment, the four coordinates of the bounding box of each shape cluster obtained in the first stage are used as clustering features. The horizontal and vertical coordinates are clustered separately using the same two-stage approach as for resizing. The average coordinates of each cluster are then determined as the final layout result. The specific threshold parameters for the Canopy algorithm will be described in more detail in Section 6.1.

\section{\textbf{Evaluation}}
To evaluate and analyze the performance of our methods, we trained and optimized the model on our \emph{hdFlowmind} dataset. A detailed description about evaluation setup, experimental contents, results, and analysis will be provided in this section.

\subsection{\textbf{Evaluation Setup}}
This section will provide an in-depth explanation of the evaluation setup for our implementation, including the metrics and baseline used to assess the performance of our model. Researchers can access our code demo on GitHub\footnote{https://github.com/cai-jianfeng/flowmind2digital.git}.

\subsubsection*{\textbf{Implementation}}
Our neural network is based on the framework of Detectron2, which utilizes Mask-RCNN and heat map keypoints detection with pytorch. For our experiments, we utilized the Keypoint-ResNet-50-FPN11 backbone, which is relatively fast and balances speed and accuracy effectively. We initialize the model weights using the pre-trained model from the COCO dataset, obtained from the Detectron2 model zoo. When it comes to hyperparameters, we mostly follow the default Detectron2 configuration for training. Specifically, the top-k anchor in train and test are 1500 and 1000 respectively in RPN, the base learning rate is 0.02, the smooth $\beta$ in $l_1$ is 0.5. Please refer to \href {https://github.com/facebookresearch/detectron2/blob/main/configs/COCO-Keypoints/keypoint_rcnn_R_50_FPN_1x.yaml}{Detectron2 configs} for more hyperparameters and architecture settings. For gradient descent, we use Adam with 80k iterations and a batch size of 4, allowing the model to see 320k augmented images (80k batches of size 4). This process takes approximately 8 hours on a GeForce RTX 3090 with 16GB memory. During training, the basic learning rate is set to the default value of 0.00025 for Adam's adaptive change of Detection2.

For post-processing, we utilize Baidu's offline service paddleOCR\footnote{https://www.paddlepaddle.org/}, specifically the PP-OCRv3 version, which supports both Chinese and English languages. This service is based on the PaddlePaddle framework and prioritizes precision and speed balance. To achieve this, it employs model slimming and depth optimization techniques. Since it is deployed in an offline environment, users can choose whether or not to perform character recognition. For automatic typesetting, we use the T1 and T2 parameters of Canopy, as shown in Table \ref{tab:T4}. For Kmeans clustering, we use the default parameters of the sklearn module.

\begin{table}[htbp]
\centering
\caption{Clustering parameters}
\begin{tabular}{ccc}
\toprule
\textbf{Canopy} & \textbf{T1(inch)} & \textbf{T2(inch)} \\
\midrule
\textbf{First Clustering} & 1 & $\min\frac{length^2+width^2}{1.618}$ \\
\midrule
\textbf{Second Clustering} & 0.8 & $\min \frac{length}{1.618}$,$ \min \frac{width}{1.618}$ \\
\bottomrule
\end{tabular}%
\label{tab:T4}%
\end{table}%

Another crucial aspect of our method involves performing non-maximum suppression between different classes to address the issue of excessive IoU. To assess its impact within our specific domain, we conducted an ablation study, comparing two models: one employing the default method from Detectron2, and the other utilizing an additional NMS between different classes after prediction to filter the results twice.

Furthermore, to evaluate the effect of adding single basic shapes to the training set, we conducted another ablation experiment that did not involve changing the parameters of other methods. This experiment involved comparing the changes in each metric before and after adding 485 images, as discussed in Section 4.

\subsubsection*{\textbf{Metrics}}
\textbf{Object detection:} To assess the performance of object detection, we utilize the same metrics as in the relevant sketch recognition approach[\cite{arrowrcnn}]. A bounding box is considered a true positive only if it is categorized correctly and overlaps fully with the ground-truth. We set the IoU threshold to 50\%, following previous works [\cite{arrowrcnn}][\cite{schafer2022sketch2process}]. We then use these true positives as the predicted object detection results and calculate the standard recall, precision, and F1 scores with an IoU threshold of 70\% during the calculation process. Additionally, we calculate the diagram accuracy (DA) [\cite{arrowrcnn}], which represents the proportion of images with completely correct object detection in all datasets, i.e., standard recall and precision are both 1 when the IoU threshold is 80\%, and the number of precision boxes and ground-truth boxes is equal. Since some images have zero predicted objects, the calculation of precision can result in division by zero, causing F1 and precision to return N/A for a single image. These images are ignored when calculating the average value. Similarly, when zero annotated objects are selected, F1 and recall return N/A. However, this situation does not arise in the training data, so it does not impact the calculation of diagram metrics for training and validation sets. Finally, when precision and recall are both zero, F1 is defined as 0.

\textbf{Connector Keypoints:} For the evaluation of the performance for connector recognition, we refer to the relevant work [\cite{schafer2022sketch2process}]. The detection of keypoints by neural network may have some error in the circular domain. But in fact, we are concerned about the connectivity of keypoints with shapes, that is, whether a group $(x_{from}, y_{from}, x_{to}, y_{to})$ can correctly find the connection between the shapes. Therefore, we apply the same post-processing method to find the nearest shape for train and test, which is mentioned in Section 4.2 to obtain the ground-truth connection and predicted connection of a connector. Then the standard Recall, Precision and F1 scores of the connector category are calculated in the same way as the object detection. In this context, true positive is defined as the correct recognition and full overlap of the three connector categories, where the two connected shapes have the same precision and true label.

\textbf{Character recognition:} Section 5.2 discusses the comparison between OCR for the entire image and OCR for the identified text box, which is also an essential aspect. To assess the performance of both methods under various scenarios outlined in Challenges (Section 3), we utilize the character error rate CER [\cite{sanchez2017icdar2017}], which is based on the editing distance [\cite{ukkonen1985finding}], for each text box string. However, since the dataset contains mathematical formulas and symbols, there are no complete Chinese and English character labels available. Thus, we randomly select 50 representative images from the dataset and manually annotate them to evaluate the performance.

\subsubsection*{\textbf{Baselines}}
In order to demonstrate the effectiveness of our approach, we conduct a comparative analysis with related studies. We find that BPMN with strict graphic definition, which is heavily focused on the professional domain, is not an ideal baseline. Instead, we choose Arrow-RCNN as it is similar to our work and has a hand-drawn sketch of almost all classes. Accordingly, we train and evaluate Arrow-RCNN models on the hdBPMN dataset to compare them with our \emph{Flowmind2digital} model under various scenarios outlined in Section 3. For Arrow-RCNN, we adhere to its default image augmentation methods and training parameters.

In order to test our hypothesis about the dataset, we conducted an experiment where we used the same model and replaced various datasets to evaluate the effectiveness of our hdFlowmind. We divided the experiment into two groups: one group was trained using a pre-trained model on flowmind and fine-tuned on the Handwritten-diagram-dataset, while the other group was trained without a pre-trained model. Our objective was to demonstrate the importance of our dataset and show that pre-trained models can be beneficial for training even in different domains.

\subsection{\textbf{Results}}
This section presents the evaluation results of our proposed \emph{Flowmind2digital} model and our \emph{hdFlowmind} dataset. First of all, the overall results are displayed, followed by the detailed results for each part of our model. Next, we illustrate the ablation studies. Finally, the time-memory complexity analysis and post-processing software docking are presented.

\subsubsection*{\textbf{Overall results and baselines}}
\begin{figure}[h]
\centering
\includegraphics[width=3.8in]{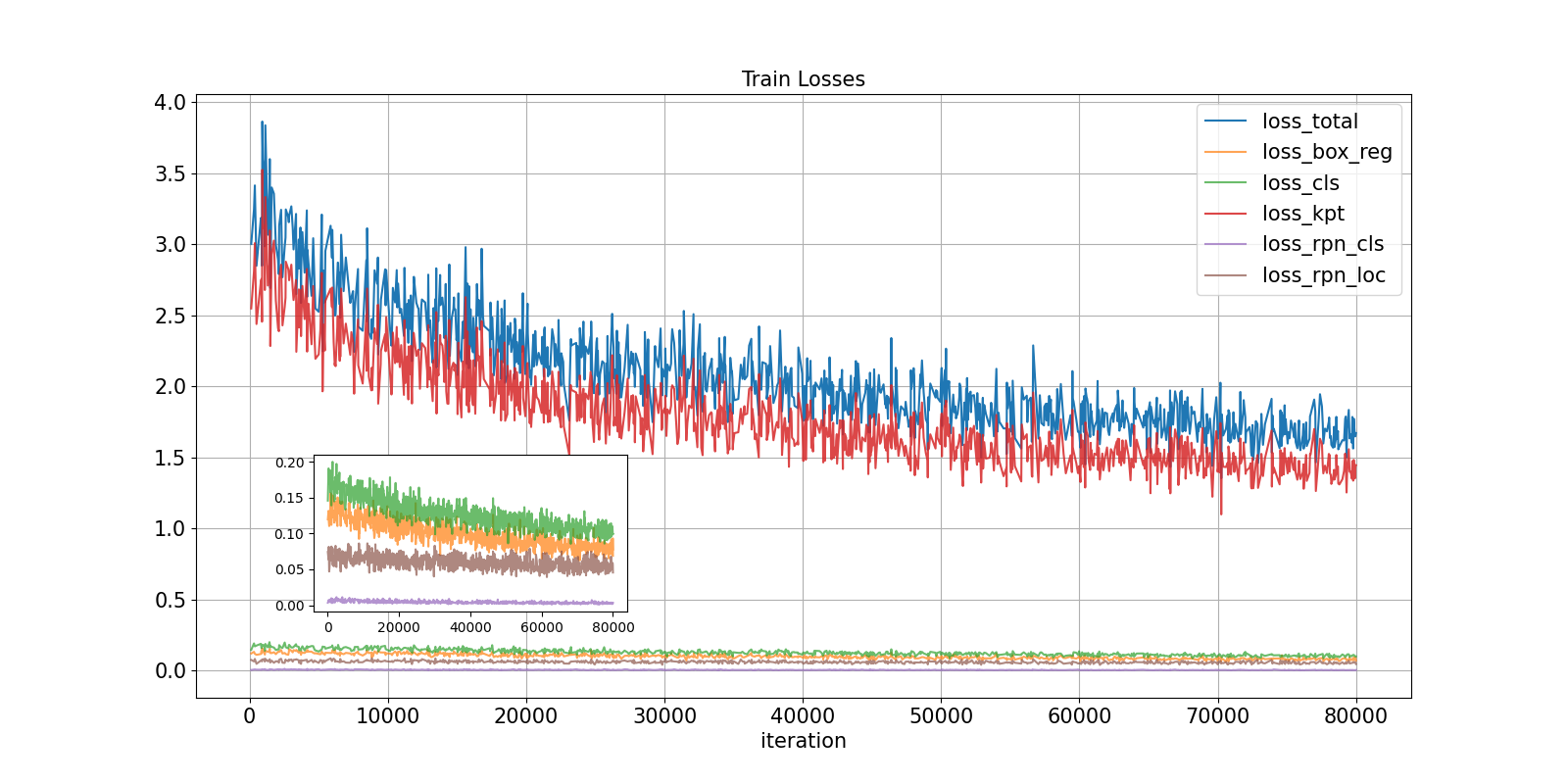}
\caption{Losses During the Whole Training Process on hdFlowmind}
\label{pho:img14}
\end{figure}
Fig. \ref{pho:img14} shows the individual loss terms and metrics throughout the 80k iterations on the \emph{hdFlowmind} dataset. It can be clearly observed that the loss mainly stems from the localization of connector keypoints, and after 80k iterations of training, the initial intense fluctuations tend to become stable. The loss of the RPN network, bounding box regression and classification is below 0.2 and gradually tends to fit as the training progresses.

The overall results and metrics of evaluation are presented in Table \ref{tab:T5} compare with Arrow-RCNN. For certain classes, the F1 score exceeds the range of Recall and Precision, due to one of them being N/A. In this case, the F1 score is also N/A and is not included in the average calculation, resulting in this outcome. Note that in the experiment, the calculation method of the four metrics is weighted average by the number of categories. As shown in Table \ref{tab:T5}, our approach has a better ability to capture complex scene features, with the total diagram accuracy that is 20\% higher than that of Arrow-RCNN, and other metrics that are about 15\% higher. Further comparison of recognition examples reveals that our method performs significantly better than Arrow-RCNN in scenarios with more noise, such as over-exposure and shadows. In terms of arrow recognition, Arrow-RCNN has poor performance in identifying multiple arrows or intersecting arrows. In terms of shape and text recognition, it struggles to distinguish styles with overlapping strokes. Overall, our method demonstrates stronger robustness and better performance in fine-grained recognition.

\begin{table*}[htbp]
\centering
\caption{Overall approach results for the test set}
\scalebox{0.8}{
\setlength{\tabcolsep}{1.2mm}{
\begin{tabular}{cccccccccccccc}
\toprule
& \multicolumn{3}{c}{Shape} & \multicolumn{3}{c}{Connector} & \multicolumn{3}{c}{Text} & \multicolumn{4}{c}{Total} \\
\midrule
\textbf{Approach} & Rec. & Prec. & F1 & Rec. & Prec. & F1 & Rec. & Prec. & F1 & Rec. & Prec. & F1 & Diagram accuracy \\
\cmidrule{2-14}
\textbf{Arrow R-CNN [\cite{arrowrcnn}]} & 0.7875 & 0.8250 & 0.8057 & 0.7342 & 0.6964 & 0.7147 & 0.7789 & 0.7472 & 0.7627 & 0.764 & 0.718 & 0.754 & 0.2\% \\
\cmidrule{2-14}
\textbf{Flowmind2digital} & 0.9698 & 0.9544 & 0.9760 & 0.8085 & 0.7577 & 0.8136 & 0.8472 & 0.8247 & 0.8407 & 0.885 & 0.865 & 0.873 & 22.5\% \\
\bottomrule
\end{tabular}}}%
\label{tab:T5}%
\end{table*}%
\begin{figure}[h]
\centering
\includegraphics[width=3.5in]{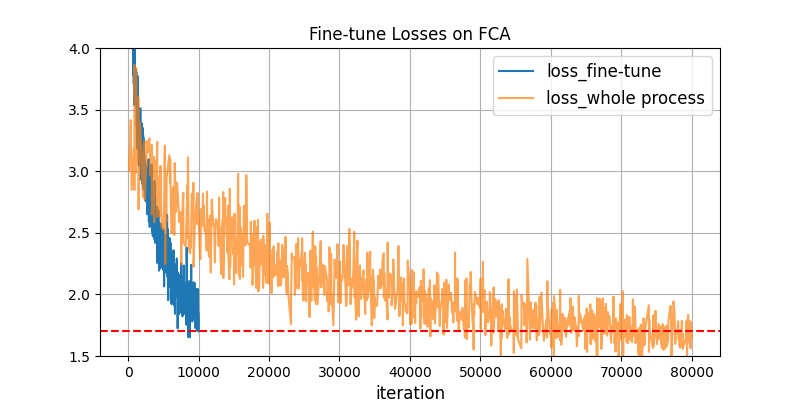}
\caption{Comparison of Training Curves with and without Fine-tuning}
\label{pho:img15}
\end{figure}

Next, we fine-tuned a pre-trained model using FC\_A, FC\_B, F\_A and hdBPMN datasets on our hdFlowmind dataset. The performance of the fine-tuned models was compared with the models trained directly on these datasets without pre-training. The model architecture and parameters used in the training process were identical to those used in the training of \emph{Flowmind2digital}. The iterations for both experimental groups was set to 5,000. From training process (Fig. \ref{pho:img15}), it can be inferred that using the \emph{hdFlowmind} pre-trained model for fine-tuning on the FC\_A dataset greatly accelerates gradient descent and loss convergence compared to not using a pre-trained model. Subsequently, we present a comparison experiment (Table \ref{tab:T6}) for fine-tuning on the aforementioned dataset. After conducting multiple experiments and taking the average results, it was found that using the pre-trained model provides a certain degree of improvement in F1 score and diagram accuracy, especially for the FCA and FCB datasets, which are both process diagrams, with an improvement of up to 6\%. For the FA dataset in the finite automata domain, due to the small number of categories, it quickly fitted with only 5,000 iterations, and the pre-trained model improved the metrics by approximately 1-2\%. However, for the BPMN scenario with strict symbol standards, the 5,000 iterations were clearly insufficient for the model converge, and the hdBPMN2021 dataset's lack of text categories resulted in only a 1-2\% improvement in recognition accuracy with the pre-trained model. Overall, our pre-trained model exhibits good generalization in the sketch domain, fully demonstrating the richness and diversity of the scenarios in our dataset.

\begin{table}[htbp]
\centering
\caption{Comparison of metric results for fine-tuning experiment on different test set of our model}
\begin{tabular}{ccccc}
\toprule
\multirow{2}[4]{*}{\textbf{Dataset}} & \multicolumn{2}{c}{\textbf{Pre-train on \emph{hdFlowmind}}} & \multicolumn{2}{c}{\textbf{No fine-tuning}} \\
\cmidrule{2-5} & F1 & DA & F1 & DA \\
\midrule
FA & 0.9492 & 0.2143 & 0.9351 & 0.1305 \\
\midrule
FCA & 0.8352 & 0.0585 & 0.8064 & 0.0350 \\
\midrule
FCB & 0.9066 & 0.1326 & 0.8425 & 0.0408 \\
\midrule
hdBPMN & 0.4347 & - & 0.4251 & - \\
\midrule
\multicolumn{5}{c}{*DA=Diagram Accuracy} \\
\bottomrule
\end{tabular}%
\label{tab:T6}%
\end{table}%

\subsubsection*{\textbf{Object detection}}
\begin{table}[htbp]
\centering
\caption{Object detection results per class obtained for the test set of our model}
\scalebox{0.9}{
\begin{tabular}{cccccc}
\toprule
\textbf{Class} & \textbf{Rec.} & \textbf{Prec.} & \textbf{F1} & \textbf{DA} & \textbf{Count}\\
\midrule
Circle & 0.979 & 1.000 & 0.986 & 0.935 & 209 \\
\midrule
Diamond & 0.985 & 0.978 & 0.980 & 0.940 & 96 \\
\midrule
Long oval & 0.992 & 0.980 & 0.981 & 0.933 & 97 \\
\midrule
Hexagon & 0.994 & 1.000 & 0.996 & 0.949 & 111 \\
\midrule
Parallelogram & 0.959 & 0.991 & 0.969 & 0.893 & 122 \\
\midrule
Rectangle & 0.986 & 0.961 & 0.968 & 0.720 & 465 \\
\midrule
Trapezoid & 0.977 & 0.970 & 0.971 & 0.940 & 71 \\
\midrule
Triangle & 0.981 & 0.976 & 0.974 & 0.902 & 127 \\
\midrule
WA & 0.9824 & 0.97807 & 0.976 & 0.8525 & 1298 \\
\midrule
\multicolumn{6}{c}{*DA = Diagram accuracy, WA = Weighted average} \\
\bottomrule
\end{tabular}}%
\label{tab:T7}%
\end{table}%
\begin{figure}[h]
\centering
\includegraphics[width=2.5in]{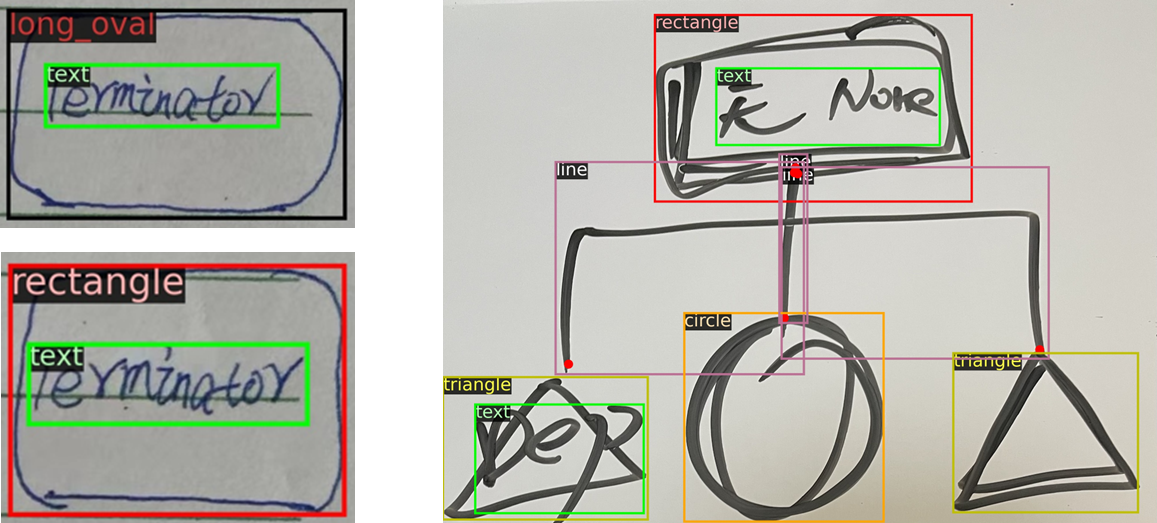}
\caption{Left: Shapes that are extremely easy for humans to confuse (long ovals and rounded rectangles). Right: The recognition result effectively solves the challenges mentioned earlier.}
\label{pho:img16}
\end{figure}

Table \ref{tab:T7} reports a detailed performance of our object detection classes in a manner that describes the weighted average metrics of each shape. The results of our experiments indicate that our model is capable of accurately recognizing the majority of the primary shapes with high accuracy, achieving a recall rate, precision rate, and F1 score around 95\%. For certain categories, due to their low occurrence in the test set, the aforementioned issues related to F1 calculation surpass recall and precision rates, and thus, further evaluation is needed.

Post-hoc analysis of the results reveals that our target detector performs excellent in different backgrounds and handwriting scenarios, and it can handle the difficulties mentioned in Section 3, such as overlapping circles, crossing, and text overprinting. However, the most challenging task is to correctly distinguish certain categories, especially the confusion between rounded rectangles and long ovals, as depicted in the Fig. \ref{pho:img16}. This is not surprising, given the subjectivity of the drawings, and recognizing these differences between hand-drawn models is also a difficult task for humans, especially in terms of the variations in curvature.

\subsubsection*{\textbf{Connector recognition}}
\begin{table}[htbp]
\centering
\caption{Connector and keypoints detection results per class obtained for the test set of our model}
\scalebox{0.9}{
\begin{tabular}{cccccc}
\toprule
\textbf{Class} & \textbf{Rec.} & \textbf{Prec.} & \textbf{F1} & \textbf{DA } & \textbf{Count} \\
\midrule
Arrow & 0.848 & 0.836 & 0.836 & 0.559 & 611 \\
\midrule
Line & 0.855 & 0.845 & 0.845 & 0.714 & 157 \\
\midrule
Double arrow & 0.750 & 0.744 & 0.743 & 0.525 & 264 \\
\midrule
WA & 0.8240 & 0.8138 & 0.8136 & 0.5739 & 1032 \\
\midrule
\multicolumn{6}{c}{*DA = Diagram accuracy, WA = Weighted average} \\
\bottomrule
\end{tabular}}%
\label{tab:T8}%
\end{table}%
Table \ref{tab:T8} further demonstrates the superiority of our model and its versatility in post-processing determination of connection as a whole. According to the first three metrics in the experimental results, arrows and lines are more easily recognized than double arrows, but the diagram accuracy is skewed due to the low number of line samples in the test set. The detection performance of double-arrows is not as good as the other two, which can be attributed to the triangular lines at the arrowhead that often overlap with shapes or are perceived as part of other shapes. Furthermore, the double arrow possesses features of both arrow and line, which increases the difficulty of recognition. This provides insight into future improvement efforts.

\begin{figure}[htbp]
\centering
\includegraphics[width=3.2in]{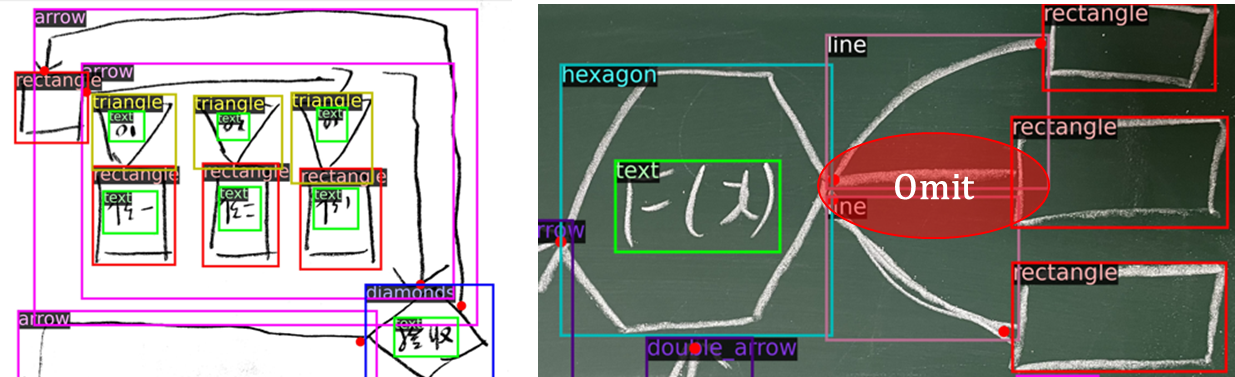}
\caption{Left: Successfully recognized large-span arrows and various fine-grained graphics. Right: Missed the short connecting line selected by the red oval.}
\label{pho:img17}
\end{figure}

Upon further analysis, our recognizer is able to handle various challenges such as many-to-many  connectors that cross the entire diagram with relatively good performance. However, for connections that are very close to each other, our model demonstrates limited ability to handle and may treat additional strokes as part of the error or as combinations of other shapes, as shown in Fig. \ref{pho:img17}. A reasonable explanation for this phenomenon lies in the intricacies of proximity-based spatial relationships. When connectors are very close, the model might face difficulties in precisely separating them, as the visual features become more challenging to differentiate. Additionally, the proximity of strokes may introduce ambiguity, leading to potential errors in the recognition process.

In conclusion, our method consistently achieves metrics approaching 80\% on connectors, highlighting its efficacy in handling intricate visual relationships, while the recognition performance for connectors may be comparatively lower than that of shapes owing to their inherent flexibility and diversity. This underscores the distinct advantages of our approach, particularly in addressing the complexities of scenarios outlined in Section 3, such as crossings.

\subsubsection*{\textbf{Textbox recognition}}
\begin{table}[htbp]
\centering
\caption{Comparison of using textbox selection and directly using OCR}
\begin{tabular}{ccccc}
\toprule
\textbf{Class} & \textbf{Rec.} & \textbf{Prec.} & \textbf{F1} & \textbf{Diagram accuracy} \\
\cmidrule{2-5}
Text box & 0.8509 & 0.8354 & 0.8407 & 0.2913 \\
\midrule
\textbf{Metric} & \multicolumn{2}{c}{\textbf{Specified region}} & \multicolumn{2}{c}{\textbf{Whole region}} \\
\cmidrule{2-5}
\textbf{CER} & \multicolumn{2}{c}{8.5\%} & \multicolumn{2}{c}{35.7\%} \\
\bottomrule
\end{tabular}%
\label{tab:T9}%
\end{table}%
As shown in Table \ref{tab:T9}, the results of employing the Baidu PaddleOCR service for handwriting recognition are poor. As demonstrated in Section 3, the main reason of low accuracy is the inability to distinguish between handwritten text and graphical intersections. However, this is not the case for our method, as the first stage of regions of interest identification directly frames the scope. At this point, OCR is used to decode the text in the specified region, and CER has seen a significant improvement, performing well on isolated text boxes. This means that if the OCR prediction is accurate and can reach the level of humans, the detected text block (region of interest) will also be accurate.

It should be noted that our method only achieves a recognition accuracy of 29\% for text boxes in images, particularly for the recognition of mathematical symbols, brackets, operators, etc. This warrants further investigation.

In conclusion, the majority of errors directly or indirectly incurred during the handwritten text block recognition process are due to OCR service errors, and fine-tuning of the ROI plays a significant role. However, this does not indicate that existing OCR services have major flaws, because high accuracy can still be achieved for pure text recognition in small areas.

\subsubsection*{\textbf{Ablation study}}
The results of our ablation study in Table \ref{tab:T10} show the benefits of adding images of single basic shapes to the training set, improving the Recall (from 0.792 to 0.889), Precision (from 0.784 to 0.872) and F1-score (from 0.786 to 0.879) to increase by 10\%. In particular, it greatly improves the diagram accuracy over 20\%. From the training process, these images greatly alleviates the over-fitting situation during model training, especially for connectors. It is, therefore, not necessary to excessively pursue sophisticated images when collecting datasets, and single target recognition is also promotional for machine learning.

\begin{table}[htbp]
\centering
\caption{Ablation study conducted on the validation set}
\scalebox{0.9}{
\setlength{\tabcolsep}{0.5mm}{
\begin{tabular}{ccccc}
\toprule
\textbf{Method} & \textbf{Rec.} & \textbf{Prec.} & \textbf{F1} & \textbf{Diagram accuracy} \\
\midrule
NMS & 0.792 & 0.784 & 0.786 & 0.066 \\
\cmidrule{2-5}
Add basic images & 0.894 & 0.864 & 0.877 & 0.275 \\
\cmidrule{2-5}
NMS+Add basic images & 0.889 & 0.872 & 0.879 & 0.283 \\
\bottomrule
\end{tabular}}}%
\label{tab:T10}%
\end{table}%

\begin{figure}[h]
\centering
\includegraphics[width=3.1in]{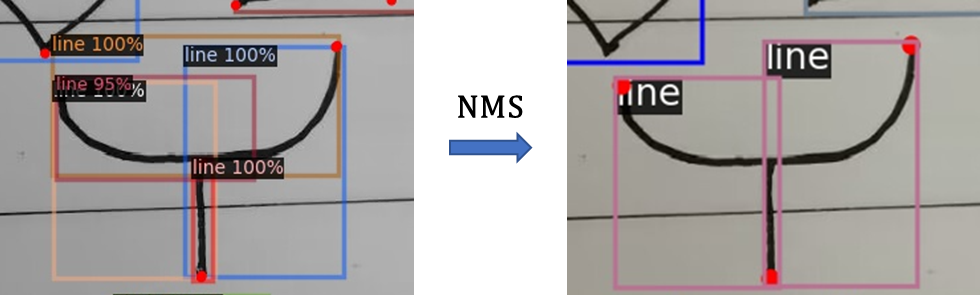}
\caption{Non-maximum suppression between different categories effectively solves the problem of many-to-many connectors being recognized multiple times and overlapping.}
\label{pho:img18}
\end{figure}

The results of applying NMS between different classes is not as effective as the former. However, it is obvious from Table \ref{tab:T10} whether NMS between different classes improves Precision-score and reduces Recall-score around 0.01. Further, this also improves the diagram accuracy. This experimental result is consistent with our intuition. NMS reduces the number of redundant bounding boxes, and also removes true positive bounding boxes, leading to the decline of recall rate, and vice versa, resulting in the improvement of precision. In the field of sketch, because of the simple shape and the arbitrariness of the user, it is common for the same basic strokes to have two categories of high accuracy at the same time. In the case of multiple connectors, additional detected lines are often generated due to crossing and masking. Therefore, in the domain of the flowmind, this assumption works effectively. Fig. \ref{pho:img18} shows an example of how this method works on sketches.

\subsubsection*{\textbf{Running time and memory}}
\begin{figure}[h]
\centering
\includegraphics[width=3.1in]{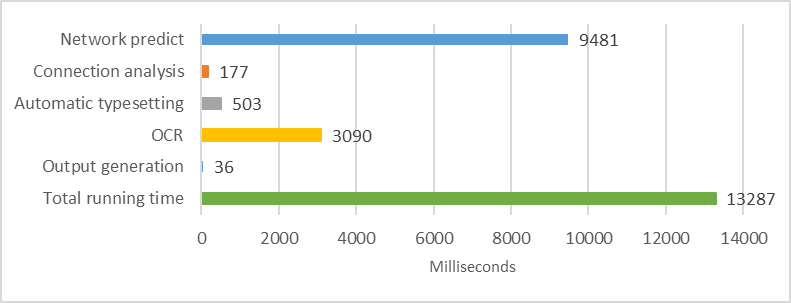}
\caption{Median runtime measures obtained for the validation set}
\label{pho:img19}
\end{figure}
Fig. \ref{pho:img19} illustrates the runtime of the components of \emph{Flowmind2digital}. Given the inconvenience of installing the Conda environment for actual users, we tested using a CPU (Interl(R) Core(TM) i5-8300H CPU @ 2.30GHz). Given the image to be processed (3000*4000 pixels), it first goes through the network prediction module (9481ms), followed by the relation analysis and graph construction (177ms), automatic typesetting (503ms), and the optional OCR module (3090ms), and finally the integrated output (36ms). The whole process took 13287ms, with most of the time spent on the two neural modules. The observed processing time of approximately 10 seconds represents a favorable trade-off between accuracy and speed on a standard CPU. This duration is considered a well-balanced solution for practical applications, aligning with the need to efficiently process and interpret visual information without compromising on detection precision. If we use GPU (e.g. GeForce RTX 3090) in inference, the whole process can be completed within one second. 

\begin{figure}[h]
\centering
\includegraphics[width=3.5in]{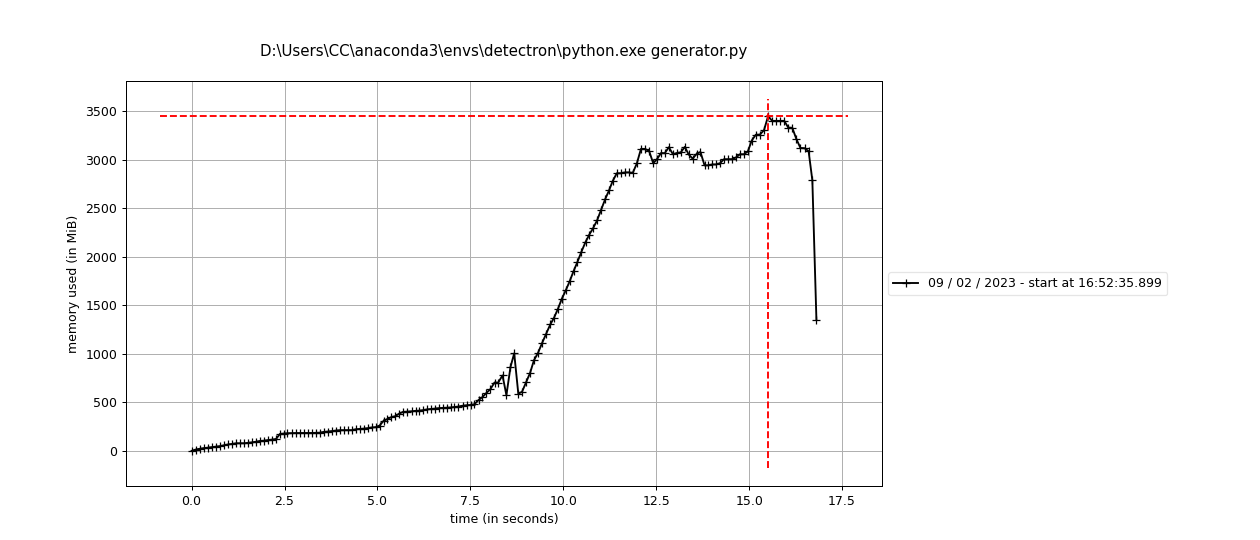}
\caption{Memory Analysis of the Whole Process}
\label{pho:img20}
\end{figure}

Fig. \ref{pho:img20} shows the memory usage of \emph{Flowmind2digital} over time. In monitoring memory usage, it is noteworthy that the initialization time of Python-related modules extends the overall processing time beyond the 10-second mark. The peak memory is around 3500MB at 15.3s, which occurs during the final integration of the module outputs. Most of the memory consumption still occurs in the neural network module, in line with the time consumption. 

\subsubsection*{\textbf{Software Results Display}}
\begin{figure*}[h]
\centering
\includegraphics[width=6.5in]{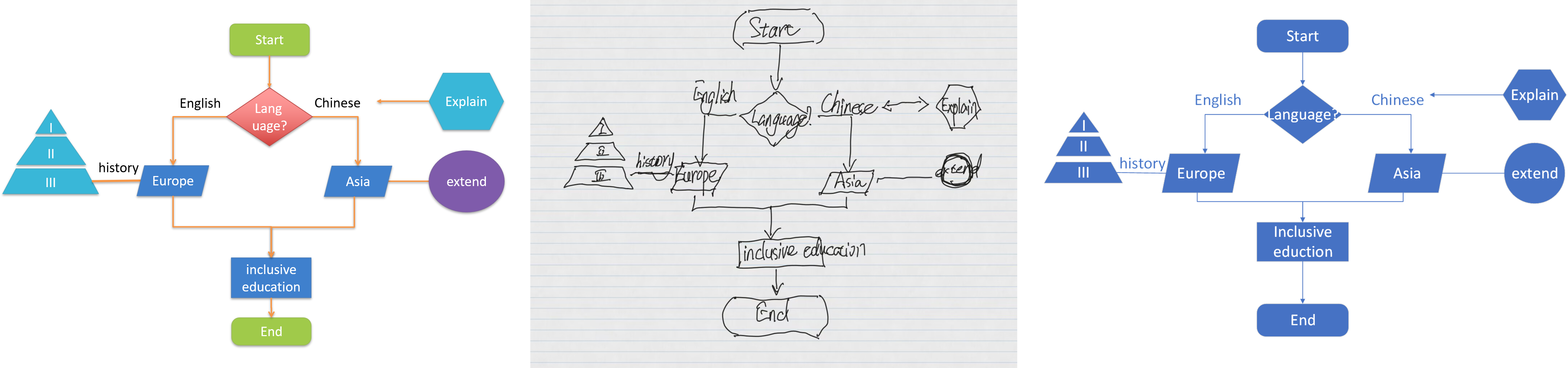}
\caption{Visualization of the Final Software Results}
\label{img21}
\end{figure*}

Finally, we display the final result of the integration of the user-inputted raw sketch image with the PPT/Visio software. Note that, based on the flexibility and standardization of both software, our result can only serve as a preliminary reference for further refinement by the designer. In PPT, the shape color is based on the principle of category consistency, while the texture and outline filling follow the default template. In Visio, we initially design the interface in blue as the main color scheme. The font and line width adapt to the size of the input graphic and the bounding boxes. The output visualization is shown in Fig. \ref{img21}.
\section{\textbf{Discussion}}
In this section, we discuss the implications, limitations, and provide reference insights for future research work.

\textbf{\emph{Implications}}. Our work extends from the specific domain of hand-drawn diagrams, such as BPMN, UML, and flowcharts, and generalizes further by considering the characteristics of natural hand-drawn diagrams. We combine the multi-connection features of mind maps with the geometric shapes that are commonly used and creatively varied in composition to better reflect the natural hand-drawn features.

Moreover, our approach establishes a more novel, convenient, and effective comprehensive processing mode. It enables brainstorming and creation on a whiteboard or even a blank sheet of paper, followed by direct digital acquisition, collection, and storage of composition information. Connection analysis and automatic layout are then performed, and finally, a document is generated using OCR services. The entire process is fully automated and intelligent. It is worth noting that because the OCR service is independent, the text recognition performance can be further improved as existing natural language processing models evolve in the future.

In addition, we have created the \emph{hdFlowmind} dataset, which is the first object and keypoint detection dataset covering both mind maps and flowcharts. It contains thousands of images and tens of thousands of annotations, and has a wide range of applications. It can effectively address the problems mentioned in the challenge. Furthermore, we found that adding simple basic shapes to the dataset can effectively solve the overfitting problem during the training process for hand-drawn sketches (without stroke sequence information), which contain far less information than RGB images.

\textbf{\emph{Limitations}}. Objectively speaking, our model also has a series of limitations. Firstly, our post-processing relies on specific software interfaces, e.g. if Microsoft PPT/Visio is no longer used as the most common presentation software, the post-processing work will need to be further modified for integration. To expand the model's application scope and integration into standardized BPMN scenarios, professional visualization software needs to be integrated. Even when the software does not provide ready-to-use documentation, library functions need to be written at the operating system level.

Secondly, although we believe that our dataset scenarios have high external validity and cover a variety of application scenarios, the images in the dataset are mainly collected from various scenarios in university life, and the overall image features and quality differ significantly. Therefore, in practical use, people may still encounter more complex situations, such as multiple people using pens of various colors to draw on the same whiteboard in a messy way.

Finally, limited by the diversity of connector paths, our model can only recognize the connection relationships and cannot distinguish between the forms of a path, such as straight or curved arrows. Moreover, the electronic version of the connector also depends on the template of the relevant visualization software and cannot truly fit the path of the hand-drawn arrow. This is also a major bottleneck in all sketch recognition works and requires further research in the future. At the same time, different expressions of the same geometric figure is also be a big challenge, which can enrich our graphic material library, such as isosceles triangle, right triangle, etc.

\section{\textbf{Conclusion}}
In this paper, we focus on the recognition problem of hand-drawn sketches. We combine the characteristics of flowcharts and mind maps, and propose our \emph{Flowmind2digital} method on the basis of existing solutions. By introducing OCR and integrating with Microsoft Power Point/Visio visualization software, it is the first comprehensive, fully automated sketch recognition method. To enrich application scenarios and better fit the characteristics of hand-drawn sketches by natural persons, we created the "\emph{hdFlowmind}" dataset, which consists of 1776 images and tens of thousands of annotations, solving the recognition difficulties encountered in actual use due to messy sketches. In the experimental evaluation process, we not only demonstrated the effectiveness of the \emph{hdFlowmind} dataset, but also showed that \emph{Flowmind2digital} is very accurate, versatile in use scenarios, capable of handling fine-grained recognition problems in complex sketches, and consistently outperforms existing algorithms.

Next, we have identified several directions for future work. First and foremost, increasing recognition accuracy and speed are undoubtedly the most importance. On the one hand, we can continue to enrich the breadth of the dataset, covering a wider variety of scenarios and increasing the number of shape categories. On the other hand, for connector recognition, research can be conducted based on its path characteristics. In text recognition, further collaboration with the NLP field will be pursued to handle more detailed information such as rotation angles of handwritten characters. Second, exploring the combination of the flowmind pre-trained model with existing research in other symbol recognition fields (such as BPMN) is also an interesting topic. The merging of research in various fields of offline sketch recognition will help deepen our understanding of the universality of human sketches (e.g., connection relationships). Finally, increasing the practicality of \emph{Flowmind2digital} in practice is also necessary. In the future, it can be inherited by more software in various fields, providing a variety of intelligent visualization solutions.

\bibliographystyle{cas-model2-names}
\nocite{*}
\bibliography{texbib2}
\end{document}